\documentclass[twoside]{article}

\usepackage[accepted]{aistats2019}

\usepackage{microtype}
\usepackage{graphicx}
\usepackage{subfig}
\usepackage{wrapfig}
\usepackage{booktabs} 
\usepackage{listings}
\usepackage{dblfloatfix}
\usepackage{float}
\usepackage{amsfonts}
\usepackage{amssymb}
\usepackage{bbm}
\usepackage{mathrsfs}

\usepackage{hyperref}

\usepackage{amsmath}
\usepackage{amsthm}
\usepackage{amssymb}
\usepackage{algorithm}
\usepackage{algpseudocode}
\usepackage{proof}
\usepackage{todonotes}
\usepackage{multicol}
\usepackage[color]{changebar}

\newtheorem{theorem}{Theorem}

\newtheorem{defn}{Definition}

\usepackage{listings}
\usepackage{color}
\usepackage{xcolor}
\definecolor{blue}{rgb}{0,0.3,0.7}
\definecolor{red}{rgb}{0.60,0.0,0.0}
\definecolor{purple}{rgb}{0.5,0,0.7}
\definecolor{cyan}{rgb}{0.0,0.6,0.5}
\definecolor{gray}{rgb}{0.4,0.4,0.4}

\lstdefinelanguage{scheme}
{sensitive, %
 alsoletter={:,-,+,*,?,/,!,>,<}, %
 morecomment=[l]{;}, %
}[comments]

\lstdefinelanguage{anglican}%
{% empty (strings and keywords)
 morekeywords=[1]{},
 morekeywords=[2]{%
   def, def-, defn, defn-, defmacro, defmulti, defmethod, %
   defstruct, defonce, declare, definline, definterface, %
   defprotocol, defrecord, defstruct, deftype, defproject, ns, %
 }, %
 morekeywords=[3]{->, ->>, .., amap, and, areduce, as->, assert, binding, %
   bound-fn, case, comment, cond, cond->, cond->>, condp, declare, definline, %
   definterface, defmacro, defmethod, defmulti, defn, defn-, defonce, %
   defprotocol, defrecord, defstruct, deftype, delay, doseq, dosync, dotimes, %
   doto, extend-protocol, extend-type, fn, for, future, gen-class, %
   gen-interface, if, if-let, if-not, if-some, import, io!, lazy-cat, lazy-seq, let, %
   letfn, locking, loop, memfn, ns, or, proxy, proxy-super, pvalues, %
   recur, refer-clojure, reify, some->, some->>, sync, time, when, when-first, %
   when-let, when-not, when-some, while, with-bindings, with-in-str, %
   with-loading-context, with-local-vars, with-open, with-out-str, %
   with-precision, with-redefs, else}, %
  morekeywords=[4]{*, *', +, +', -, -', ->ArrayChunk, ->Vec, ->VecNode, %
    ->VecSeq, -cache-protocol-fn, -reset-methods, /, <, <=, =, ==, >, >=, %
    accessor, aclone, add-classpath, add-watch, agent, agent-error, %
    agent-errors, aget, alength, alias, all-ns, alter, alter-meta!, %
    alter-var-root, ancestors, apply, array-map, aset, aset-boolean, aset-byte, %
    aset-char, aset-double, aset-float, aset-int, aset-long, aset-short, assoc, %
    assoc!, assoc-in, associative?, atom, await, await-for, await1, bases, bean, %
    bigdec, bigint, biginteger, bit-and, bit-and-not, bit-clear, bit-flip, %
    bit-not, bit-or, bit-set, bit-shift-left, bit-shift-right, bit-test, %
    bit-xor, boolean, boolean-array, booleans, bound-fn*, bound?, butlast, byte, %
    byte-array, bytes, cast, char, char-array, char?, chars, chunk, %
    chunk-append, chunk-buffer, chunk-cons, chunk-first, chunk-next, chunk-rest, %
    chunked-seq?, class, class?, clear-agent-errors, clojure-version, coll?, %
    commute, comp, comparator, compare, compare-and-set!, compile, complement, %
    concat, conj, conj!, cons, constantly, construct-proxy, contains?, count, %
    counted?, create-ns, create-struct, cycle, dec, dec', decimal?, delay?, %
    deliver, denominator, deref, derive, descendants, destructure, disj, disj!, %
    dissoc, dissoc!, distinct, distinct?, doall, dorun, double, double-array, %
    doubles, drop, drop-last, drop-while, empty, empty?, ensure, %
    enumeration-seq, error-handler, error-mode, eval, even?, every-pred, every?, %
    ex-data, ex-info, extend, extenders, extends?, false?, ffirst, file-seq, %
    filter, filter-ns-publics, filterv, find, find-keyword, find-ns, %
    find-protocol-impl, find-protocol-method, find-var, first, flatten, float, %
    float-array, float?, floats, flush, fn?, fnext, fnil, force, format, %
    frequencies, future-call, future-cancel, future-cancelled?, future-done?, %
    future?, gensym, get, get-in, get-method, get-proxy-class, %
    get-thread-bindings, get-validator, group-by, hash, hash-combine, hash-map, %
    hash-ordered-coll, hash-set, hash-unordered-coll, identical?, identity, %
    ifn?, in-ns, inc, inc', init-proxy, instance?, int, int-array, integer?, %
    interleave, intern, interpose, into, into-array, ints, isa?, iterate, %
    iterator-seq, juxt, keep, keep-indexed, key, keys, keyword, keyword?, last, %
    line-seq, list, list*, list?, load, load-file, load-reader, load-string, %
    loaded-libs, long, long-array, longs, macroexpand, macroexpand-1, %
    make-array, make-hierarchy, map, map-indexed, map?, mapcat, mapv, max, %
    max-key, memoize, merge, merge-with, meta, method-sig, methods, min, %
    min-key, mix-collection-hash, mod, munge, name, namespace, namespace-munge, %
    neg?, newline, next, nfirst, nil?, nnext, not, not-any?, not-empty, %
    not-every?, not=, ns-aliases, ns-functions, ns-imports, ns-interns, %
    ns-macros, ns-map, ns-name, ns-publics, ns-refers, ns-resolve, ns-unalias, %
    ns-unmap, nth, nthnext, nthrest, num, number?, numerator, object-array, %
    odd?, parents, partial, partition, partition-all, partition-by, pcalls, %
    peek, persistent!, pmap, pop, pop!, pop-thread-bindings, pos?, pr, pr-str, %
    prefer-method, prefers, print, print-ctor, print-simple, print-str, printf, %
    println, println-str, prn, prn-str, promise, proxy-call-with-super, %
    proxy-mappings, proxy-name, push-thread-bindings, quot, rand, rand-int, %
    rand-nth, range, ratio?, rational?, rationalize, re-find, re-groups, %
    re-matcher, re-matches, re-pattern, re-seq, read, read-line, read-string, %
    realized?, record?, reduce, reduce-kv, reduced, reduced?, reductions, ref, %
    ref-history-count, ref-max-history, ref-min-history, ref-set, refer, %
    release-pending-sends, rem, remove, remove-all-methods, remove-method, %
    remove-ns, remove-watch, repeat, repeatedly, replace, replicate, require, %
    reset!, reset-meta!, resolve, rest, restart-agent, resultset-seq, reverse, %
    reversible?, rseq, rsubseq, satisfies?, second, select-keys, send, send-off, %
    send-via, seq, seq?, seque, sequence, sequential?, set, %
    set-agent-send-executor!, set-agent-send-off-executor!, set-error-handler!, %
    set-error-mode!, set-validator!, set?, short, short-array, shorts, shuffle, %
    shutdown-agents, slurp, some, some-fn, some?, sort, sort-by, sorted-map, %
    sorted-map-by, sorted-set, sorted-set-by, sorted?, special-symbol?, spit, %
    split-at, split-with, str, string?, struct, struct-map, subs, subseq, %
    subvec, supers, swap!, symbol, symbol?, take, take-last, take-nth, %
    take-while, test, the-ns, thread-bound?, to-array, to-array-2d, trampoline, %
    transient, tree-seq, true?, type, unchecked-add, unchecked-add-int, %
    unchecked-byte, unchecked-char, unchecked-dec, unchecked-dec-int, %
    unchecked-divide-int, unchecked-double, unchecked-float, unchecked-inc, %
    unchecked-inc-int, unchecked-int, unchecked-long, unchecked-multiply, %
    unchecked-multiply-int, unchecked-negate, unchecked-negate-int, %
    unchecked-remainder-int, unchecked-short, unchecked-subtract, %
    unchecked-subtract-int, underive, unsigned-bit-shift-right, update-in, %
    update-proxy, use, val, vals, var-get, var-set, var?, vary-meta, vec, %
    vector, vector-of, vector?, with-bindings*, with-meta, with-redefs-fn, %
    xml-seq, zero?, zipmap}, %
  morekeywords=[5]{def-cps-fn, defanglican, defm, defquery, defun, defproc, defdist}, %
  morekeywords=[6]{cps-fn, fm, lambda, query, with-primitive-procedures}, %
  morekeywords=[7]{%
    doquery, %
    conditional, %
    collect-by, equalize, exec, infer, log-marginal, print-predicts, %
    rand, rand-int, rand-nth, rand-roulette, stripdown, warmup, %
    ->CRP-process, ->DP-process, ->GP-process, %
    ->bernoulli-distribution, ->beta-distribution, ->binomial-distribution, %
    ->categorical-crp-distribution, ->categorical-distribution, %
    ->categorical-dp-distribution, ->chi-squared-distribution, %
    ->dirichlet-distribution, ->discrete-distribution, %
    ->exponential-distribution, ->flip-distribution, ->gamma-distribution, %
    ->mvn-distribution, ->normal-distribution, ->poisson-distribution, ->factor-distribution, %
    ->sample, ->observe, sample*, observe*, %
    ->uniform-continuous-distribution, ->uniform-discrete-distribution, %
    ->wishart-distribution, CRP, DP, GP, abs, absorb, acos, asin, atan, %
    bernoulli, beta, binomial, categorical, categorical-crp, categorical-dp, %
    cbrt, ceil, chi-squared, cos, cosh, cov, dirichlet, discrete, exp, %
    exponential, flip, floor, gamma, gen-matrix, log, log-gamma-fn, %
    log-mv-gamma-fn, log-sum-exp, map->CRP-process, map->DP-process, %
    map->GP-process, map->bernoulli-distribution, map->beta-distribution, %
    map->binomial-distribution, map->categorical-crp-distribution, %
    map->categorical-distribution, map->categorical-dp-distribution, %
    map->chi-squared-distribution, map->dirichlet-distribution, %
    map->discrete-distribution, map->exponential-distribution, %
    map->flip-distribution, map->gamma-distribution, map->mvn-distribution, %
    map->normal-distribution, map->poisson-distribution, %
    map->factor-distribution,
    map->uniform-continuous-distribution, map->uniform-discrete-distribution, %
    map->wishart-distribution, mvn, normal, factor, poisson, pow, produce, %
    rint, round, signum, sin, sinh, sqrt, tag, tan, tanh, transform-sample, %
    uniform-continuous, uniform-discrete, wishart, uniform, %
    map->fn,op,
    add-log-weight, add-predict, clear-predicts, get-log-weight, %
    get-mem, get-predicts, in-mem?, set-log-weight, set-mem, %     
  }, %
  morekeywords=[8]{mem, observe, predict, retrieve, sample, store}, %
  sensitive, %
  alsoletter={:,-,+,*,?,/,!,>,<,.}, %
  morecomment=[l][\color{gray}]{;}, %
  morestring=[b]", %
}[keywords,comments,strings]
 
\lstdefinestyle{default}{language=Anglican,basicstyle=\ttfamily\small, columns=flexible, showstringspaces=false}
\lstdefinestyle{clojure}{language=Anglican,basicstyle=\ttfamily\small, columns=flexible, showstringspaces=false}

\usepackage{lipsum}

\usepackage{titlesec}
\titlespacing\section{0pt}{6pt plus 2pt minus 2pt}{2pt plus 2pt minus 0pt}
\titlespacing\subsection{0pt}{6pt plus 2pt minus 2pt}{2pt plus 2pt minus 0pt}
\titlespacing\subsubsection{0pt}{6pt plus 2pt minus 2pt}{2pt plus 2pt minus 0pt}

\begin{document}

\runningtitle{LF-PPL}

\runningauthor{Yuan Zhou, Bradley J.~Gram-Hansen, Tobias Kohn, Tom Rainforth, Hongseok Yang, Frank Wood}
\twocolumn[
\aistatstitle{LF-PPL: A Low-Level First Order Probabilistic Programming Language for Non-Differentiable Models}

\aistatsauthor{ Yuan Zhou\textsuperscript{*,1} \And Bradley J.~Gram-Hansen\textsuperscript{*,1}
	\And  Tobias Kohn\textsuperscript{2,$\dag$} 	\And Tom Rainforth\textsuperscript{1} 
}
\aistatsauthor{
	Hongseok Yang\textsuperscript{3} \And Frank Wood\textsuperscript{4}}

\aistatsaddress{ \textsuperscript{1}University of Oxford~  \textsuperscript{2}University of Cambridge~  \textsuperscript{3}KAIST~   \textsuperscript{4}University of British Columbia}
]
\begin{abstract} 
	\vspace{-6pt}
% !TEX root = main_lfppl.tex

We develop a new Low-level, First-order Probabilistic Programming Language~(LF-PPL) suited for
models containing a mix of continuous, discrete, and/or piecewise-continuous variables.
The key success of this language and its compilation scheme is in its ability to 
automatically distinguish parameters the density function is discontinuous with respect to,
 while further providing runtime checks for boundary crossings. 
This enables the introduction of new inference engines that are able to 
exploit gradient information, while remaining efficient for models which are not everywhere differentiable.
We demonstrate this ability by incorporating a discontinuous Hamiltonian Monte Carlo (DHMC) inference
engine that is able to deliver automated and efficient inference for non-differentiable models.
Our system is backed up by a mathematical formalism that ensures that any model expressed in this language
has a density with measure zero discontinuities to maintain the validity of the inference engine.

\end{abstract}

\setlength{\abovedisplayskip}{3.5pt}
\setlength{\belowdisplayskip}{3.5pt}
\setlength{\abovedisplayshortskip}{3.5pt}
\setlength{\belowdisplayshortskip}{3.5pt}	

\section{Introduction}
\label{sec:intro}
% !TEX root = main_lfppl.tex

Non-differentiable densities arise in a huge variety of common probabilistic
models~\cite{mohasel2016probabilistic, gelman2013bayesian}.
Often, but not exclusively, they
occur due to the presence of discrete variables.
In the context of probabilistic programming~\cite{gordon2014probabilistic,goodman2012church, wood2014new,gelman2015stan} 
such densities are
often induced via branching, i.e. \texttt{if-else} statements, where the predicates depend upon the latent variables of the model.
Unfortunately, performing efficient and scalable inference in models with non-differentiable densities 
is difficult and algorithms adapted for such problems typically require specific knowledge about the
discontinuities~\cite{afshar2015reflection,nishimura2017discontinuous,lee2018reparameterization}, such as
which variables the target density is discontinuous with respect to and catching occurrences of the sampler 
crossing a discontinuity boundary.
However, detecting when discontinuities occur is difficult and problem dependent. 
Consequently, automating specialized inference algorithms 
in probabilistic programming languages (PPLs) is challenging.

To address this problem, we introduce a new Low-level First-order Probabilistic Programming Language~(LF-PPL),
with a novel accompanying compilation scheme.
Our language is based around carefully chosen mathematical constraints, such that the set of discontinuities 
in the density function of any model written in LF-PPL will have measure zero.
This is an essential property for many 
inference algorithms designed for non-differentiable densities~\cite{afshar2015reflection,nishimura2017discontinuous,lee2018reparameterization,dinh2017probabilistic,yi2017roll}. 
Our accompanying compilation scheme automatically classifies discontinuous and continuous random variables for any model specified in our language.
Moreover, this scheme can be used to detect transitions across discontinuity boundaries at
runtime, providing important information for running such inference schemes.

Relative to previous languages, LF-PPL enables one to incorporate a broader class of specialized inference 
techniques as automated inference engines. In doing so, it removes the burden from the user
of manually establishing which variables the target is not differentiable with
respect to.
Its low-level nature is driven by a desire to establish the
minimum language requirements to support inference engines tailored to problems with measure-zero discontinuities,
and to allow for a formal proof of correctness. Though still usable in its own right, our main intention is 
that it will be used as a compilation target for existing systems, or as an intermediate system for
designing new languages.

There are a number of different derivative-based inference paradigms 
for which LF-PPL can help extend to non-differentiable setups~\cite{afshar2015reflection,nishimura2017discontinuous,lee2018reparameterization,dinh2017probabilistic,yi2017roll}.
Of particular note, are
stochastic variational inference (SVI)~\cite{hoffman2013stochastic,ranganath2014black,
blei2017variational,kucukelbir2015automatic} and
Hamiltonian Monte Carlo (HMC)~\cite{duane1987hybrid,neal2011mcmc},
two of the most widely used approaches for probabilistic programming 
inference.

In the context of the former,
\cite{lee2018reparameterization} recently showed that the
reparameterization trick can be generalized to piecewise differentiable models when the 
non-differentiable boundaries can be identified, leading to an approach which provides
significant improvements over previous methods that do not explicitly account for the discontinuities.
LF-PPL provides a framework that could be used to apply their approach in a probabilistic programming
setting, thereby paving the way for significant performance improvements for such models.

Similarly,
many variants of HMC have been proposed  in recent years
to improve the sample efficiency and scalability when the target density is non-differentiable~\cite{afshar2015reflection,nishimura2017discontinuous,zhang2012continuous, pakman2013auxiliary, pakman2014exact}.
Despite this, no probabilitic programming systems (PPSs) support these tailored approaches at present, as the underlying languages are not able to extract the necessary information for their automation.
The novel compilation approach of LF-PPL provides key information
for running such approaches, enabling their implementation as automated inference engines.
 We realize this potential by
implementing Discontinuous HMC (DHMC)~\cite{nishimura2017discontinuous}
as an inference engine in LF-PPL, allowing for efficient, automated,
HMC-based inference in models with a mixture of continuous and discontinuous variables.

\section{Background and Related Work}
\label{sec:motivation}
% !TEX root = main_lfppl.tex

There exists a number of different approaches to probabilistic programming
that are built around a variety of semantics and inference engines.
Of particular relevance to our work are PPSs designed around derivative
based inference methods that exploit automatic differentiation~\cite{baydin2017automatic}, such as Stan~\cite{gelman2015stan}, PyMC3~\cite{salvatier2016probabilistic}, Edward~\cite{tran2017deep}, Turing~\cite{turing} and Pyro~\cite{Pyro}.
Derivative based inference algorithms have been an essential component in
enabling these systems to provide efficient and, in particular, scalable
inference, permitting both large datasets and high dimensional models.

One important challenge for these systems occurs in
dealing with probabilistic programs that contain discontinuous densities and/or variables.  
From the statistical perspective, dealing with discontinuities is often important 
for conducting effective inference.  
For example, in HMC, discontinuities can cause statistical inefficiency
by inducing large errors in the leapfrog integrator, leading to potentially very low acceptance 
rates~\cite{afshar2015reflection,nishimura2017discontinuous}.  In other words, though the leapfrog integrator
remains a valid, reversible, MCMC proposal even when discontinuities break the reversibility of
the Hamiltonian dynamics themselves, they can undermine the effectiveness of this proposal. 

Different methods have been suggested to improve inference performance in models with discontinuous densities.
For example, 
they use sophisticated integrators in the HMC setting to
remain effective when there are discontinuities.  
Analogously, in the variational
inference and deep learning literature, reparameterization methods have been proposed that
allow training for discontinuous targets and discrete 
variables~\cite{lee2018reparameterization,maddison2016concrete}.  

However, these advanced methods are, in general, not incorporated in existing gradient-based PPSs,
as existing systems do not have adequate support to deal with the discontinuities 
in the density functions of the model defined by probabilistic programs. 
This is usually necessary 
to guarantee the correct execution of those inference methods in an automated fashion,
as many require the set of discontinuities to be of measure zero.
That is, the union of all points where the density is discontinuous have zero measure 
with respect to the Lebesgue measure.
In addition to this, some further methods require knowledge of where the discontinuities are, or
at least catching occurrences of discontinuity boundaries being crossed.

Of particular relevance to our language and compilation scheme 
are compilers which compile the program to an
artifact representing a direct acyclic graphical model (DAG), such as those employed in BUGS~\cite{spiegelhalter1996bugs} and, in particular,
the first order PPL (FOPPL) explored in \cite{van2018introduction}.
Although the dependency structures of the programs in our language are established in a similar manner,
unlike these setups, programs in our language will not always correspond to a DAG, due to different restrictions on our density factors, as will be explained in the next section.
We also impose necessary constraints on the language by limiting the functions allowed to ensure that the advanced inference processes remain valid.

\section{The Language}
\label{sec:intLF-PPL}
% !TEX root = main_lfppl.tex

LF-PPL adopts a Lisp-like syntax, like that of Church \cite{goodman2012church} and Anglican \cite{wood2014new}. 
The syntax contains two key special forms,
\lstinline[style=clojure]{sample} and
\lstinline[style=clojure]{observe}, between which the
distribution of the query is defined and whose interpretation and syntax
is analogous to that of Anglican.  
  
More precisely, \lstinline[style=clojure]{sample} is
used for drawing random variables, returning that variable,
and \lstinline[style=clojure]{observe} factors the density of the program using
existing variables and fixed observations, returning \lstinline[style=clojure]{zero}.
Both special forms are designed to take a \emph{distribution object} as input,
with \lstinline[style=clojure]{observe} further taking an observed value.  
These distribution
objects form the elementary random procedures of the language and
are constructed using one of a number of internal constructors for common objects
such as \lstinline[style=clojure]{normal} and  \lstinline[style=clojure]{bernoulli}.
Figure~\ref{fig:fopplfig} shows an example of an LF-PPL program.

A distribution object constructor
of particular note is  \lstinline[style=clojure]{factor}, which
can only be used with  \lstinline[style=clojure]{observe}.
Including the statement \lstinline[style=clojure]{(observe (factor log-p) _)}
will factor the program density using the value of \lstinline[style=clojure]{(exp log-p)}, 
with no dependency  on the observed value itself (here \lstinline[style=clojure]{_}).  
The significance of \lstinline[style=clojure]{factor} is that it allows the specification
of arbitrary unnormalized target distributions, quantified as \lstinline[style=default]{log-p} which can 
be generated internally in the program, and thus have the form of any
deterministic function of the variables that can be 
written in the language.  

\begin{figure}[tbp]
\begin{lstlisting}[style=default]
(let [x (sample (uniform 0 1))]
    (if (< (- q x) 0)
        (observe (normal 1 1) y)
        (observe (normal 0 1) y))
    (< (- q x) 0))
\end{lstlisting}
\vspace{-13pt}
\caption{An example LF-PPL program sampling $x$ from a uniform random variable and invoking a choice between two \lstinline[style=clojure]{observe} statements that factor the trace weight using different Gaussian likelihoods.
The \lstinline[style=clojure]{(< (- q x) 0)} term, which is usually written as ${((q-x)<0)}$,
represents a Bernoulli variable parameterized by q and its boolean value also corresponds to which branch of the \lstinline[style=clojure]{if} statement is taken.
The slightly unusual writing of the program is due to its deliberate low-level nature, with almost all syntactic sugar removed.
One sugar that has been left in for exposition is an additional term in the let block, i.e. 
\lstinline[style=clojure]{(let [x e] e e)}, which can be trivially unraveled. }
\vspace{-16pt}
\label{fig:fopplfig}
\end{figure}

Unlike many first-order PPLs, such as that of~\cite{van2018introduction}, LF-PPL programs do not permit interpretation
as DAGs because we allow the observation of internally sampled variables and the use
of \lstinline[style=clojure]{factor}.  This increases the range of models that can be encoded and is, 
for example, critical in allowing undirected models to be written. LF-PPL programs need not 
correspond to a correctly normalized joint formed by the combination of prior and likelihood terms.
Instead we interpret the density of a program in the manner outlined by~\cite[\S4.3.2 and \S4.4.3]{rainforth2017automating},
noting that for any LF-PPL program, the number of \lstinline[style=clojure]{sample} and \lstinline[style=clojure]{observe} statements (i.e. $n_x$ and $n_y$ in their
notation) must be fixed,
a restriction that is checked during the compilation.

To formalize the syntax of LF-PPL, let us use $x$ for a real-valued variable, $c$ for a real number, \lstinline[style=clojure]{op} for an analytic primitive operation on
reals, such as \lstinline[style=clojure]{+}, \lstinline[style=clojure]{-},  \lstinline[style=clojure]{*},  \lstinline[style=clojure]{/} and \lstinline[style=clojure]{exp}, and $d$ for a distribution object
whose density is defined with respect
to a Lebesgue measure and is piecewise smooth under 
analytic partition~(See Definition~\ref{def-1}).
Then the syntax of expressions $e$ in our language are given as:
\vspace{-3pt}
\[
\begin{array}{@{}r@{}c@{}l@{}}
e & \,::=\, & 
x \mid c \mid (\text{\lstinline[style=clojure]{op}}\ e\,\ldots\,e) 
\mid (\text{\lstinline[style=clojure]{if}}\ (< e\ 0)\ e\ e)
\mid (\text{\lstinline[style=clojure]{let}}\ [x\ e]\ e)
\\
& \,\,\,\,\,\, \mid & (\text{\lstinline[style=clojure]{sample}}\ (d\ e\, \ldots\, e))
\mid (\text{\lstinline[style=clojure]{observe}}\ (d\ e\, \ldots\, e)\ c)

\vspace{-5pt}
\end{array}
\]

Our syntax is deliberately low-level to permit theoretical
analysis and aid the exposition of the compiler.  
However, common syntactic sugar such as \texttt{for}-loops and higher-level
branching statements can be trivially included using straightforward
unravellings.  Similarly, we can permit discrete variable distribution objects by
noting that these can themselves be desugared to a combination of continuous
random variables and branching statements.
Thus, it is straightforward to extend this minimalistic framework
to a more user-friendly language using standard compilation approaches,
such that LF-PPL will form an intermediate representation.  
For implementation and code, see \url{https://github.com/bradleygramhansen/PyLFPPL}.

\vspace{-2pt}
\section{Compilation Scheme}
\label{sec:compilation-overall}
% !TEX root = main_lfppl.tex

We now provide a high-level 
description of how the compilation process works.
Specifically, we will show how it transforms an arbitrary LF-PPL program to a representation 
that can be exploited by an inference engine that makes of use of discontinuity information.

The compilation scheme  performs three core tasks:  a) finding the  
variables which the target is discontinuous with respect to,
b) extracting the density of the program to a convenient form that can be used by an inference engine, and c) allowing boundary crossings to be detected at runtime.
Key to providing these features is the construction of an internal representation of the program that specifies the dependency structure of the variables, the \textit{Linearized Intermediate Representation} (LIR). The LIR contains vertices, arc pairs,
and information of the \lstinline[style=clojure]{if} predicates.
Each vertex of the LIR denotes a \lstinline[style=clojure]{sample} 
or \lstinline[style=clojure]{observe} statement, of which only a finite 
and fix number can occur in LF-PPL.
The arcs of the LIR define both the probabilistic and \lstinline[style=clojure]{if} condition
dependencies of the variables. 
The former of these are constructed in same was as is done in the FOPPL compiler detailed in~\cite{van2018introduction}.

Using the dependency structure represented by the LIR, we can establish which variables are capable of changing the path taken by a program trace, that is the change the branch taken by one or more \lstinline[style=clojure]{if} statements.
Because discontinuities only occur in LF-PPL through \lstinline[style=clojure]{if} statements, the target must be continuous with respect to any variables not capable of changing the traversed path.  We can thus mark these variables as being ``continuous''.
Though it is possible for the target to still be continuous with respect to variables that appear in, or have dependent variables appearing in, the branching function of an \lstinline[style=clojure]{if} statement, such cases cannot, in general, be statically established.  
We therefore mark all such variables  as ``discontinuous''.

To extract the density to a convenient form for the inference engine, the compiler transforms the program into a collection four sets---$\Delta,\Gamma,D,$ and $F$---by recursively applying the translation rules given in Section \ref{sec:translation-rule}.
Here $\Delta$ specifies the set of all variables sampled in the program, while $\Gamma$ specifies only the variables marked as discontinuous. 
$D$ represents the density associated with all the \lstinline[style=clojure]{sample} statements in a program, while $F$ represents the density factors originating for the \lstinline[style=clojure]{observe} statements, along with information on the program return value.  
These densities are themselves represented through a collection of smooth density terms and indicator functions truncating them into disjoint regions, each corresponding to a particular program path.  This construction will be discussed in depth in Section~\ref{sec:translation-rule}.

To catch boundary crossings at run time, each \lstinline[style=clojure]{if}  
predicate is assigned a unique boolean variable within the LIR.
We refer to these variables as \emph{branching variables}.
The boolean value of the branching variable denotes whether the current sample falls into the \lstinline[style=clojure]{true} or \lstinline[style=clojure]{false} branch of the corresponding  \lstinline[style=clojure]{if} statement and is used to signal boundary crossings at runtime.
Specifically, if one branching variable changes its boolean value, 
this indicates that at least one sampled variables effecting that \lstinline[style=clojure]{if} predicate has crossed the boundary. 
The inference engine can therefore track changes in the set of all Boolean values to catch the boundary crossings.

We finish the section by noting two limitations of the compiler and for discontinuity detection more generally.
 Firstly, we note that it is possible to construct programs which have piecewise smooth densities that contain regions of zero density.
Though it is important to allow this ability, for example to construct truncated distributions, it may cause issues for certain inference algorithms if it causes the target to have disconnected regions of non-zero density.
As analytic densities are either zero everywhere or ``almost-nowhere'' (see Section~\ref{sec:theory}), we (informally) have that all realizations of a program that take a particular path will either have zero density or all have a non-zero density.
Consequently, it is relatively straight forward to establish if a program has regions of zero density.  
However,  whether these regions lead to ``gaps'' is far more challenging, and potentially impossible, to establish.
Moreover, constructing inference procedures for such problems is extremely challenging.  
We therefore do not attempt to tackle this issue in the current work.

A second limitation is that changes in the vector of branching variables is only a \emph{sufficient} condition for the occurrence of a boundary crossing.
This is because it is possible for \emph{multiple} boundaries to be crossed in a single update that results in the new sample following the same path as the old one.
For example, when moving from $x=-0.5$ to $x=1.5$ then a branching variable corresponding to $x^3-x>0$ returns \lstinline[style=clojure]{true} in both cases even though we have crossed two boundaries.
The problem of establishing with certainty that \emph{no} boundaries have been crossed when moving between two points is mathematically intractable in the general case.
As this problem is not specific to the probabilistic programming setting, we do not give it further consideration here, noting only that it is important from the perspective of designing inference algorithms that convergence is not undermined by such occurrences.

\section{Mathematical Foundation and Compilation Details}
\label{sec:theory-all}
% !TEX root = main_lfppl.tex
Our story so far was developed by introducing a low-level first-order probabilistic programming 
language (LF-PPL) and its accompanying compilation scheme. 
We shall now expose the underlying mathematical details, 
which ensure that discontinuities contained within the densities of the programs one can compile in LF-PPL are of a suitable measure. 
This enables us to satisfy the requirements of several inference algorithms for non-differentiable densities. 
We also provide the formal translation rules of the LF-PPL, which are built around these mathematical underpinnings.

\subsection{Piecewise Smooth Functions}
\label{sec:theory}
A function $\mathcal{G} : \mathbb{R}^k \to \mathbb{R}$ is \emph{analytic} if it
is infinitely differentiable and its multivariate Taylor expansion at any point
$x_0 \in \mathbb{R}^k$ absolutely converges to $\mathcal{G}$ point-wise in a neighborhood of $x_0$.
Most primitive functions that we encounter 
in machine learning and statistics are analytic, and the composition of analytic functions is also analytic.

\begin{defn}\label{defn:piecewise-smooth-analytic}
	A function $\mathcal{G} : \mathbb{R}^k \to \mathbb{R}$ is \emph{piecewise smooth
		under analytic partition} if it has the following form:
	\vspace{-10pt}
	\[
	\mathcal{G}(x)
	=
	\sum_{i = 1}^N
	\left(
	\prod_{j = 1}^{M_i} \mathbbm{1}[p_{i,j}(x) \geq 0]
	\cdot
	\prod_{l = 1}^{O_i} \mathbbm{1}[q_{i,l}(x) < 0]
	\cdot
	h_i(x)
	\right)
	\]
	\vspace{-10pt}
	where
	\begin{enumerate}
		\item the $p_{i,j}, q_{i,l} : \mathbb{R}^k \to \mathbb{R}$ are analytic;
		\vspace{-5pt}\item the $h_i : \mathbb{R}^k \to \mathbb{R}$ are smooth;
		\vspace{-5pt}\item $N$ is a positive integer or $\infty$;
		\vspace{-5pt}\item $M_i, O_i$ are non-negative integers; and
		\vspace{-5pt}\item the indicator functions
		\vspace{-5pt}
		\[
		\prod_{j = 1}^{M_i} \mathbbm{1}[p_{i,j}(x) \geq 0] \cdot \prod_{l = 1}^{O_i} \mathbbm{1}[q_{i,l}(x) < 0]
		\vspace{-5pt}
		\]
		for the indices $i$ define a partition of $\mathbb{R}^k$, that is,
		the following family forms a partition of $\mathbb{R}^k$:
		\vspace{-5pt}\[
		\bigg\{
		\left\{x \,{\in}\, \mathbb{R}^k
		\;\Big|\;
		\begin{array}{@{}l@{}}
		\forall j \, p_{i,j}(x) \,{\geq}\, 0,\,\forall l \, q_{i,l}(x) \,{<}\, 0
		\end{array}\right\}
		\Big|\, 1 \,{\leq}\, i \,{\leq}\, N\bigg\}.
		\]
	\end{enumerate}
\label{def-1}
\end{defn}
\vspace{-15pt}
Intuitively, $\mathcal{G}$ is a function defined by partitioning
$\mathbb{R}^k$ into finitely or countably many regions
and using a smooth function $h_i$ within region $i$.
The products of the indicator functions of these summands form a partition of $\mathbb{R}^k$,
so that only one of these products gets evaluated to a non-zero value
at $x$. 
To evaluate the sum, we just need to evaluate
these products at $x$ one-by-one until we find one
that returns a non-zero value. 
Then, we have to compute
the function $h_i$ corresponding to this product at the
input $x$.
Even though the number of
summands (regions) $N$ in the definition is countably infinite, we can
still compute the sum at a given $x$. 
%Note that when $N = 1$ and $M_1 = O_1 = 0$, there is no partition in the space. 

\begin{theorem}
	\label{defn:theorem-1}
	If an unnormalized density $\mathcal{P} : \mathbb{R}^{n} \to \mathbb{R}_+$
	has the form of Definition~\ref{defn:piecewise-smooth-analytic} and so is piecewise smooth under analytic partition,
	then there exists a (Borel) measurable subset $A \subseteq \mathbb{R}^{n}$ such
	that $\mathcal{P}$ is differentiable outside of $A$ and the Lebesgue measure of $A$ is zero.	
\end{theorem}
\vspace{-5pt}
The proof is given in Appendix A. The target density being almost everywhere differentiable with discontinuities of measure zero is an important property required by many inference techniques for non-differentiable models~\cite{nishimura2017discontinuous}.
As we shall prove in Section~\ref{sec:translation-rule}, 
any program that can be compiled in LF-PPL constructs a density in the form of Definition~\ref{defn:piecewise-smooth-analytic}, and thus satisfies this necessary condition. 

\subsection{Translation Rules}
\label{sec:translation-rule}
% !TEX root = main_lfppl.tex
\subsubsection{Overview}
The compilation scheme $e \leadsto (\Delta,\Gamma,D,F)$ translates
a program, which can be denoted as an expression $e$ according to the syntax in Section~\ref{sec:intLF-PPL}, 
to a quadruple of sets $(\Delta,\Gamma,D,F)$. 
The first set $\Delta$ represents the set of all sampled random variables.
All variables generated from \lstinline[style=clojure]{sample} statements in $e$ will be recognized and stored in $\Delta$. 
Variables that have not occurred in any \lstinline[style=clojure]{if} predicate are guaranteed to be continuous. 
Otherwise, they will be also put in $\Gamma \subseteq \Delta$, as the overall density is discontinuous with respect to them.
$D$ represents the densities from \lstinline[style=clojure]{sample} statements 
and has the form of a set of the pairs, i.e. 
$D = \{(\eta_1,k_1), \dots, (\eta_{N_D},k_{N_D})\}$,
where $N_D$ is the number of the pairs, $\eta$ denotes a product of indicator functions indicating the partition of the space, 
and $k$ represents the products of the densities defined by the \lstinline[style=clojure]{sample} statements.
The last set $F$ contains the densities from \lstinline[style=clojure]{observe} statements and the return expression of $e$.
It is a set of tuples 
$F = \{(\zeta_1,l_1, v_1), \dots, (\zeta_{N_F},l_{N_F}, v_{N_F}) \}$,
where  $N_F$ is the number of the tuples,
$\zeta$ functions similar to $\eta$,
$l$ is the product of the densities defined by \lstinline[style=clojure]{observe} statements 
and $v$ denotes the returning expression. 
Note that it is a design choice to have $v$ included in $F$.

Given $e \leadsto (\Delta,\Gamma,D,F)$, 
one can then construct the unnormalized density defined by the program $e$ as
\vspace{-2pt}
\begin{align}
\label{eq:pdf}
\mathcal{P} \, {:=} 
\Big( \sum_{i = 1}^{N_D} \eta_i {\cdot} k_i \Big) \cdot \Big( \sum_{j = 1}^{N_F} \zeta_j {\cdot} l_j \Big)
\end{align}
which by Theorem~\ref{theorem-2} will be piecewise smooth under analytic partitions.

Recall that by assumption, the density of each distribution type $d$ is piecewise smooth
under analytic partition when viewed as a function of a sampled value and its parameters. 
Thus, we can assume that the probability density of a distribution has the form in Definition~\ref{defn:piecewise-smooth-analytic}.
For each distribution $d$, we define a set of pairs 
$\Phi^{(d)} {=} \{(\psi_1, \phi_1), \dots, (\psi_{N_{\Phi}}, \phi_{N_{\Phi}}) \}$
where $N_{\Phi}$ is the number of the partitions,
$\psi$ denotes the product of indicator functions indicating the partition of the space,
taking the form of $\prod_{j = 1}^{M_i} \mathbbm{1}[p_{i,j}(\mathbf{x}) {\geq} 0] \cdot \prod_{l = 1}^{O_i} \mathbbm{1}[q_{i,l}(\mathbf{x}) {<} 0]$,
and $\phi$ represents a smooth probability density function within that partition.
One can then construct the probability density function $\mathcal{P}_d$ for $d$ from $\Phi^{(d)}$.
For given parameters $x_1,\ldots,x_{s}$ of the distribution $d$ and a given \lstinline[style=clojure]{sample} value $x_0$, we let $\mathbf{x} = (x_0,\ldots,x_s)$ 
and the probability density function defined by $d$ is,
\vspace{-3pt}
\begin{align*}
\mathcal{P}_d(x_0; x_1,\ldots,x_s) = {\sum}_{n = 1}^{N_{\Phi}} \psi_n(\mathbf{x}) \cdot \phi_n(\mathbf{x})
\vspace{-8pt}
\end{align*}
For example, given $x_0$ drawn from normal distribution $\mathcal{N}(\mu, \sigma)$, we have
 $\Phi^{(d)} = \left\{ \left( 1, \mathcal{N}(x_0 \, ; \mu,\sigma)\right) \right\}$ and 
 $\mathcal{P}_d(x_0; \mu,\sigma) = \mathcal{N}(x_0 \, ; \mu,\sigma)$.
Similarly a uniform $\mathcal{U}(a,b)$ sampled variable $x_0$ has  $\Phi^{(d)}$ as
\vspace{-3pt}
\begin{align*}
\big\{ &\left( \mathbbm{1}[x_0{-}a {<} 0], \, 0 \right), \ \left( \mathbbm{1}[b{-}x_0 {<} 0], \, 0 \right),\\
\vspace{-5pt}
&\left( \mathbbm{1}[x_0{-}a {\geq} 0] {\cdot} \mathbbm{1}[b{-}x_0 {\geq} 0], \, \mathcal{U}(x_0; a, b)  \right) \big\},
\vspace{-10pt}
\end{align*}
and  
$\mathcal{P}_{d} = \mathbbm{1}[x_0{-}a {\geq} 0] {\cdot} \mathbbm{1}[b{-}x_0 {\geq} 0] \cdot \mathcal{U}(x_0; a, b)$.
Note that in practice one can omit the pair $(\psi_n, \phi_n)$ in $\Phi^{(d)}$ when $\phi_n{=}\,0$ for simplicity and 
the probability density in the region denoting by the corresponding $\psi_n$ is zero.

\subsubsection{Formal Translation Rules}
The translation process $e \leadsto (\Delta, \Gamma, D, F)$,
is defined recursively on the structure of $e$. We present this
recursive definition using the following notation
\[
\infer{
	\mbox{conclusion}
}{
	\mbox{premise}
}
\vspace{-5pt}
\]
which says that if the premise holds, then the conclusion holds too.
Also, for real-valued functions $f(x_1,\ldots,x_n)$ and $f'(x_1,\ldots,x_n)$
on real-valued inputs, we write $f[x_i := f']$ to denote the composition
$
f(x_1,\ldots,x_{i-1},f'(x_1,\ldots,x_n),x_{i+1},\ldots,x_n).
$
We now define the formal translation rules.

The first two rules define how we map the set of variables $x$ and the set of constants $c$,
to their unnormalized density and the values at which they are evaluated.
\[
\begin{array}{@{}c@{}}
\infer{
	x \leadsto (\{x\},\emptyset,\,\{(1,1)\},\,\{(1,1,x)\})
}{}
\\[2ex]
\infer{
	c \leadsto (\emptyset,\emptyset,\,\{(1,1)\},\,\{(1,1,c)\})
}{}
\end{array}
\]
The third rule allows one to translate the primitive operations \lstinline[style=clojure]{op} defined in the LF-PPL, 
such as \lstinline|+|, \lstinline|-|, \lstinline|*| and \lstinline|/| with their argument expressions $e_1$ to $e_n$, 
where  $e_1$ to $e_n$ will be evaluated first. 
Note that $(\eta_i,k_i) \in D_i$ represents the enumeration of all $(\eta_i,k_i)$ pairs in $D_i$ and the result of this operation among all the $D_i$ is the possible combination of all their elements. 
For example, given three sets $D_1$, $D_2$ and $D_3$ which have three, one and two pairs respectively as their elements, the result set $D'$ will have six pairs.
This notation holds to the rest of the paper.
\[
\begin{array}{@{}c@{}}
\infer{
	(\text{\lstinline[style=clojure]{op}}\ e_1\ \ldots\ e_n) \leadsto (\bigcup_{i=1}^n \Delta_i, \bigcup_{i = 1}^{n}\Gamma_i,\,D',\,F')
}{
	e_i \leadsto (\Delta_i,\Gamma_i, \, D_i, \, F_i) \ \mbox{for $1 \leq i \leq n$} \hfill
	\\[0.5ex]
	D' = \{(\prod_{i=1}^n \eta_i,\, \prod_{i=1}^n k_i) \mid(\eta_i,k_i) \in D_i \}
	\hfill
	\\[0.5ex]
	F' = \{(\prod_{i=1}^n \zeta_i,\,\prod_{i=1}^n l_i,\, \text{\lstinline[style=clojure]{op}}\, (v_1,\ldots,v_n)) \mid (\zeta_i,l_i,v_i) \in F_i\}
}
\vspace{6pt}
\end{array}
\] 
The fourth rule for control flow operation \text{\lstinline[style=clojure]{if}} enables us to translate the
predicate $(<\,e_1\ 0)$, its consequent $e_2$ and alternative $e_3$. 
This provides us with the semantics to correctly construct a piecewise smooth function, 
that can be evaluated at each of the partitions.
\[
\begin{array}{@{}c@{}}
\infer{
	(\text{\lstinline[style=clojure]{if}}\ (<\,e_1\ 0)\ e_2\ e_3)
	\leadsto
	(\bigcup_{i=1}^3 \Delta_i, \Delta_1 \cup \Gamma_2 \cup \Gamma_3 ,\, D',\, F')
}{
	e_i \leadsto (\Delta_i,\Gamma_{i}, D_i, F_i)
	\ \mbox{for $i = 1,2,3$}
	\hfill
	\\[0.5ex]
	D' = \{(\prod_{i=1}^3 \eta_i,\,\prod_{i=1}^3 k_i) \mid
	(\eta_i,k_i) \in D_i\}
	\hfill
	\\[0.5ex]
	F' = \{(\zeta_1\cdot \zeta_2 \cdot \mathbbm{1}[v_1 < 0],\,l_1\cdot l_2,\,v_2),\hfill
	\\[0.5ex]
	\phantom{F' = \{}(\zeta_1\cdot \zeta_3 \cdot \mathbbm{1}[v_1\geq 0],\,l_1\cdot l_3,\,v_3) \mid 
	(\zeta_i,l_i,v_i) \in F_i\}
	\hfill
}
\vspace{6pt}
\end{array}
\]
The translation rule for the \lstinline[style=clojure]{sample} statement generates a random variable from a specific distribution. 
During translation, we pick a fresh variable, i.e. a variable with a unique name
to represent
this random variable and add it to the $\Delta$ set. Then we compose the density of this variable 
according to the distribution $d$ and corresponding parameters $e_i$. 
\[
\begin{array}{@{}c@{}}
\infer{
	(\text{\lstinline[style=clojure]{sample}}\ (d\, e_1\,\ldots\,e_n))
	\leadsto
	(\Delta', \Gamma', D',F')
}{
	e_i \leadsto (\Delta_i,\Gamma_i, D_i, F_i)
	\ \mbox{for $i = 1,\ldots,n$}\hfill
	\\[0.5ex]
	\mbox{pick a fresh variable $z$}\hfill
	\\[0.5ex]
	\Delta' = \{z\} \cup \bigcup_{i = 1}^{n}\Delta_i,\quad
	\Gamma' = \bigcup_{i = 1}^{n}\Gamma_i\hfill
	\\[0.5ex]
	D_0 = \{(\psi {\cdot} \prod_{i=1}^n \zeta_i, \; \phi[\mathbf{x} := (z,v_1,\ldots,v_n)]) \mid {}\hfill
	\\[0.5ex]
	\qquad\qquad\qquad\qquad\qquad\qquad
	(\psi,\phi) \in \Phi^{(d)}, (\zeta_i,l_i,v_i) \in F_i\}\hfill
	\\[0.5ex]
	D' = \{(\prod_{i=0}^n \eta_i,\; \prod_{i=0}^n k_i)
	\mid (\eta_i,k_i) \in D_i\}\hfill
	\\[0.5ex]
	F' = \{(\prod_{i=1}^n \zeta_i,\; \prod_{i=1}^n l_i,\; z) \mid (\zeta_i,l_i,v_i) \in F_i\}
	\hfill
}
\vspace{6pt}
\end{array}
\]
The translation rule for the \lstinline[style=clojure]{observe} statement, different from the \lstinline[style=clojure]{sample} expression, factors the density
according to the distribution object, with all parameters $e_i$ and the observed data $c$ evaluated. 
\[
\begin{array}{@{}c@{}} 
\infer{
	(\text{\lstinline[style=clojure]{observe}}\ (d\, e_1\, \ldots\, e_n)\ c)
	\leadsto
	(\Delta', \Gamma',D', F')
}{
	e_i \leadsto (\Delta_i,\Gamma_i, D_i, F_i)
	\ \mbox{for $i = 1,\ldots,n$}\hfill
	\\[0.5ex]
	\Delta' = \bigcup_{i = 1}^{n}\Delta_i,\quad
	\Gamma' = \bigcup_{i = 1}^{n}\Gamma_i\hfill
	\\[0.5ex]
	D' = \{(\prod_{i=1}^n \eta_i,\; \prod_{i=1}^n k_i) \mid (\eta_i,k_i) \in D_i\}\hfill 
	\\[0.5ex]
	F' = \{(\psi \cdot \prod_{i=1}^n \zeta_i,\; \phi[\mathbf{x} := (c,v_1,\ldots,v_n)]{\cdot} \prod_{i=1}^n l_i, \; 0) \mid\hfill
	\\[0.5ex]
	\qquad\qquad\qquad\qquad\qquad\quad \quad
	{} (\psi,\phi) \in \Phi^{(d)}, (\zeta_i,l_i,v_i) \in F_i\}\hfill
}
\vspace{5pt}
\end{array}
\]
The translation rule for \lstinline[style=clojure]{let} expressions
first translates the definition $e_1$ of $x$ and the body $e_2$ of \lstinline[style=clojure]{let},
and then joins the results of these translations. When joining the $\Delta$ and $\Gamma$ sets,
the rule checks whether $x$ appears in the sets from the translation of $e_2$, and
if so, it replaces $x$ by variable names appearing in $e_1$, an expression
that defines $x$. Although \lstinline[style=clojure]{let} is defined as single binding, we can construct the rules to translate 
the \lstinline[style=clojure]{let} expression, defining and binding multiple variables by properly \emph{desugaring}.
\[
\begin{array}{@{}c@{}}
\infer{
	(\text{\lstinline[style=clojure]{let}}\ [x\ e_1]\ e_2) \leadsto
	(\Delta',\, \Gamma', D', F')
}{
	e_i \leadsto (\Delta_i,\Gamma_i, D_i, F_i)\ \mbox{for $i = 1,2$}
	\hfill
	\\[0.5ex]
	\Delta_0 = \{z \mid (\zeta_1,l_1,v_1) \in F_1 \mbox{ and $z$ occurs free in $v_1$}\}
	\hfill
	\\[0.5ex]
	\Delta' = \Delta_1 \cup (\Delta_2 \setminus\{x\}) \cup (\mbox{if $(x \in \Delta_2)$ then $\Delta_0$ else $\emptyset$})
	\hfill
	\\[0.5ex]
	\Gamma' = \Gamma_1 \cup (\Gamma_2\setminus\{x\}) \cup (\mbox{if $(x \in \Gamma_2)$ then $\Delta_0$ else $\emptyset$})
	\hfill
	\\[0.5ex]
	D' = \{(\zeta_1 {\cdot} \eta_1 {\cdot} \eta_2[x:=v_1],\; k_1 {\cdot} k_2[x:=v_1]) \mid {}
	\hfill
	\\[0.5ex]
	\quad\qquad\qquad\qquad\qquad\qquad (\eta_i,k_i) \in D_i, (\zeta_1,l_1,v_1) \in F_1\}
	\hfill
	\\[0.5ex]
	F' = \{(\zeta_1 {\cdot} \zeta_2[x:=v_1],\; l_1 {\cdot} l_2[x:=v_1],\; v_2[x:=v_1]) \hfill
	\\[0.5ex]
	\qquad\qquad\qquad\qquad\qquad\qquad
	{} \mid (\zeta_i,l_i,v_i) \in F_i\}\hfill
	\hfill
}
\end{array}
\]

\vspace{3pt}
\begin{theorem}
	If $e$ is an expression that does not contain any free variables and $e \leadsto (\Delta,\Gamma, \,D,\,F)$, 
	then the unnormalized density defined by $e$ is in the form of Equation~\ref{eq:pdf}. It
	is a real-valued function on the variables in $\Delta$,
	which is non-negative and piecewise smooth under analytic partition as per Definition~\ref{def-1}.
	\label{theorem-2}
\end{theorem}
The proof is provided in Appendix~\ref{sec:theorem-2-proof}. 
By providing this set of mathematical translations we have been able to prove that any such program 
written in LF-PPL constructs a density in the form of Definition~\ref{defn:piecewise-smooth-analytic},
which is piecewise smooth under analytic partitions. 
Together with Theorem~\ref{defn:theorem-1}, we further show that this density is almost everywhere differentiable and the discontinuities are of measure zero,
a necessary condition for several inference schemes such as DHMC~\cite{nishimura2017discontinuous}.

\subsection{A Compilation Example}
\label{sec:compilation-example}
\vspace{-5pt}
% !TEX root = main_lfppl.tex

We now present a simple example of how the
compiler transforms the program $e_{pp}$ in Figure~\ref{fig:fopplfig} to the quadruple $(\Delta_{pp},\Gamma_{pp},D_{pp},F_{pp})$.
The translation rules are applied recursively and within each rule, all individual components are compiled eagerly first.
Namely, we step into each individual component and step out until it is fully compiled.
A desugared version of $e_{pp}$ is:
\begin{lstlisting}[style=default]
(let [x (sample (uniform 0 1))]
     (let [x_ (if (< (- q x) 0)
                   (observe (normal 1 1) y)
                   (observe (normal 0 1) y))]
         (< (- q x) 0)))
\end{lstlisting}
\vspace{-10pt}
where $q$ and $y$ are constant and ${x\_}$ is not used. 
It follows the following steps.
\renewcommand{\theenumi}{\roman{enumi}} 
\begin{enumerate}
	\vspace{-8pt}
	\item \label{item-outlet}
	\emph{Rule $(\text{\lstinline[style=clojure]{let}}\ [x\ e_{1,out}]\ e_{2,out})$}. 
	We start by looking at the outer let expressions, with $e_{1,out}$ being the \lstinline[style=clojure]{sample} statement and
	$e_{2,out}$ corresponding to the entire inner \lstinline[style=clojure]{let} block. 
	Before we can generate the output of this rule, we step into $e_{1,out}$ and $e_{2,out}$ and compile them accordingly. 
	\vspace{-3pt}
	\item \label{item-sample}
	\emph{Rule $(\text{\lstinline[style=clojure]{sample}}\ (d\ e_1\ e_2))$.} 
	We then apply the \lstinline[style=clojure]{sample} rule on $e_{1,out} := \text{\lstinline[style=clojure]{(sample (uniform 0 1))}}$ from \ref{item-outlet}, with each of its components evaluated first. 
	For \lstinline[style=clojure]{(uniform 0 1)}, $0$ and $1$ are constant and we have
	$0 \leadsto (\emptyset, \emptyset, \{(1,1)\}, \{(1,1,0)\})$ and $1 \leadsto (\emptyset, \emptyset, \{(1,1)\}, \{(1,1,1)\})$. 
	$d$ represents \lstinline[style=clojure]{uniform} distribution and has the form
	$\Phi^{(d)} = \left\{ \left( \mathbbm{1}[x\geq 0] {\cdot} \mathbbm{1}[1{-}x\geq 0], \, \mathcal{U}(\cdot\,;0,1) \right) \right\}$.
	 After combining each set following the rule, with a fresh variable $z$, we have
	$e_{1,out} \leadsto \big(\{z\}, \emptyset, \{(\mathbbm{1}[z{\geq}0]{\cdot}\mathbbm{1}[1{-}z{\geq}0], \mathcal{U}(z;0,1))\}, \{(1, 1, z)\} \big)$.
	\item \label{item-innerlet} 	\vspace{-5pt}
	\emph{Rule $(\text{\lstinline[style=clojure]{let}}\ [x\ e_{1,in}]\ e_{2,in})$}. We now step into $e_{2,out}$ from \ref{item-outlet} with itself being a \lstinline[style=clojure]{let} expression. 
	$e_{1,in}$ is the entire \lstinline[style=clojure]{if} statement and $e_{2,in}$ is the returning value \lstinline[style=clojure]{(< (- q x) 0)}.
	Similarly, we need to compile $e_{1,in}$ and $e_{2,in}$ first before having the result for $e_{2,out}$.
	\vspace{-3pt}
	\item \label{item-if}
	\emph{Rule $(\text{\lstinline[style=clojure]{if}}\ (< e_1 \ 0) \ e_2 \ e_3)$.} 
	To apply the \lstinline[style=clojure]{if} rule on $e_{1,in}$, we again need to compile its each individual component first.
	We start with its predicate $e_1{:=}\text{\lstinline[style=clojure]{(- q x)}}$,
	which follows the rule $(\text{\lstinline[style=clojure]{op}}\ e_1\ e_2)$.
	Then $e_1 \leadsto \big(\{x\},\, \emptyset, \, \{(1, 1)\}, \{(1,1, (q-x))\}\big)$ with $(q-x)$ as a operation \lstinline[style=clojure]{-} applied to $q$ and $x$. 
	
	$e_2$ and $e_3$ both follow
	$(\text{\lstinline[style=clojure]{observe}}\ (d\ e_1\ e_2)\ c)$.
	Take $e_2 := (\text{\lstinline[style=clojure]{observe}} \ (\text{\lstinline[style=clojure]{normal}} \ 1 \ 1) \ y)$ as an example, $1$ is constant and $d$ is the \lstinline[style=clojure]{normal} distribution and has
	$\Phi^{(d)} = \left\{ \left( 1, \mathcal{N}(\cdot\,;1,1)\right) \right\}$.
	We combine each set and have
	$e_2 \leadsto \big(\emptyset, \, \emptyset, \, \{(1, 1)\}, \ \{(1, \, \mathcal{N}(y;\,1,1), \, 0)\} \big)$.
	Similarly, $e_3 \leadsto (\emptyset, \, \emptyset, \, \{(1, 1)\}, \, \{(1, \, \mathcal{N}(y;\,0,1),\, 0)\} )$.

	With $e_{1}$, $e_{2}$ and $e_{3}$ all evaluated, we can now continue the \lstinline[style=clojure]{if} rule.
	The key features are to extract variables in $e_1$ and put into $\Gamma$ and to construct the indicator functions from $e_1$ and take the densities on each branch respectively.
	As a result, $e_{1,in}$ compiles to 
	$\Delta = \{x\}$, $\ \Gamma = \{x\}$, $\ D = \{ (1, 1) \}$ and
		$F = \big\{\big(\mathbbm{1}[q{-}x{<}0], \mathcal{N}(y;1,1), 0\big), \big(\mathbbm{1}[q{-}x{\geq}0], \mathcal{N}(y;0,1), 0\big)\big\}$.
	\item \label{item-return} 	\vspace{-5pt}
	\emph{Rule $(\text{\lstinline[style=clojure]{op}}\ e_1\ \ldots\ e_n)$.} 
	For $e_{2,in}$ in \ref{item-innerlet}, \lstinline[style=clojure]{(< (- q x) 0)} compiles to
	$(\{x\},\, \emptyset, \, \{(1, 1)\}, \{(1,1, (q{-}x<0))\})$.
	\item \label{item-innerlet-result} \vspace{-5pt}
	\emph{Result of the inner \lstinline[style=clojure]{let}.} 
	Together with the outcome from {\ref{item-if}} and {\ref{item-return}},
	we can continue compiling the inner \lstinline[style=clojure]{let} block as in {\ref{item-innerlet}}, and it is translated to
	\vspace{-3pt}
	\[\begin{array}{@{}l@{}}
	\Delta = \{x\}, \ \Gamma = \{x\},\\
	D = \big\{(\mathbbm{1}[q{-}x< 0], \, 1), (\mathbbm{1}[q{-}x \geq 0], \, 1)\big\} \\
	F = \big\{(\mathbbm{1}[q{-}x<0],\, \mathcal{N}(y;\,1,\,1), (q{-}x<0)), \  \\
	\quad \quad \ \ (\mathbbm{1}[q{-}x\geq 0],\, \mathcal{N}(y;\,0,\,1), (q{-}x<0))\big\}
	\end{array}\]
	\item \emph{Result of the outer \lstinline[style=clojure]{let}.}
	Finally, with $e_{1,out}$ compiled in {\ref{item-sample}} and $e_{2,out}$ in {\ref{item-innerlet-result}}, 
	we step out to {\ref{item-outlet}}. %$(\text{\lstinline[style=clojure]{let}}\ [x\ e_{1,out}]\ e_{2,out})$.
	It is worth to emphasize that the variables $\Delta$ are the sampled ones rather than what are named in the \lstinline[style=clojure]{let} expression, i.e. $x$ and $x\_$. 
	Here $x$ is replaced by $z$ as declared in $e_{1,out}$ by following the \lstinline[style=clojure]{let} rule, and we have the final quadruple output:
	\vspace{-3pt}
	\[\begin{array}{@{}l@{}}
	\Delta_{pp} = \{z\}, \ \Gamma_{pp} = \{z\}, \\
	D_{pp} = \Big\{\big(\mathbbm{1}[z {\geq} 0]{\cdot} \mathbbm{1}[1{-}z{\geq} 0]{\cdot}\mathbbm{1}[q{-}z{<} 0], \, \mathcal{U}(z;0, 1)\big), \\
	\vspace{-3pt}
	\quad \quad \quad \  \  \big(\mathbbm{1}[z{\geq} 0]{\cdot} \mathbbm{1}[1{-}z{\geq} 0] {\cdot} \mathbbm{1}[q{-}z{\geq} 0], \, \mathcal{U}(z;0, 1)\big)\Big\} \\
	\vspace{-3pt}
	F_{pp} = \Big\{\big(\mathbbm{1}[q{-}z{<}0],\, \mathcal{N}(y;\,1,\,1), (q{-}z{<}0)\big), \  \\
	\quad \quad \quad \  \  \big(\mathbbm{1}[q{-}z{\geq} 0],\, \mathcal{N}(y;\,0,\,1), (q{-}z{<}0)\big)\Big\}
	\end{array}\]
\end{enumerate}
\vspace{-10pt}
From the quadruple, we have the overall density as 
$\mathcal{P} = \mathbbm{1}[z{\geq} 0]{\cdot} \mathbbm{1}[1{-}z{\geq} 0] {\cdot} \mathbbm{1}[q{-}z{<} 0] {\cdot}\mathcal{U}(z;0, 1){\cdot}\mathcal{N}(y;1,1) + 
\mathbbm{1}[z{\geq} 0]{\cdot} \mathbbm{1}[1{-}z{\geq} 0] {\cdot} \mathbbm{1}[q{-}z{\geq} 0] {\cdot}\mathcal{U}(z;0, 1){\cdot}\mathcal{N}(y;0,1)$.
We can also detect when any random variable in $\Gamma$, in this case $z$, has crossed the discontinuity, by checking the boolean value of the predicate of the \lstinline[style=clojure]{if} statement  \lstinline[style=clojure]{(< (- q x) 0)}, as discussed in Section~\ref{sec:compilation-overall}

\section{Example Inference Engine: DHMC}
\label{sec:dhmcLF-PPL}
We shall now demonstrate an example inference algorithm that is compatible with 
LF-PPL.  
Specifically, we provide an implementation of
discontinuous HMC~(DHMC)\cite{nishimura2017discontinuous},
a variant of HMC for performing statistically efficient inference on probabilistic models 
with non-differentiable densities, using LF-PPL as a compilation target.
This satisfies the necessary requirement of DHMC
that the target density being piecewise smooth with discontinuities of measure zero.
Given the quadruple output from LF-PPL, DHMC updates variables in $\Gamma$ by the coordinate-wise integrator and the rest of the variables in $\Delta{\setminus}\Gamma$ by the standard leapfrog integrator.
In an existing PPS without a special support,
the user would be required to manually specify all the discontinuous 
and continuous variables, in addition to implementing DHMC accordingly.
See Appendix~\ref{sec:supp-dhmc} for further details.

\vspace{-2pt}
\subsection{Gaussian Mixture Model~(GMM)}
\vspace{-3pt}
% !TEX root = main_lfppl.tex

\begin{figure}[t]
	\centering
	\vspace{-5pt}
	\includegraphics[width=0.5\linewidth]{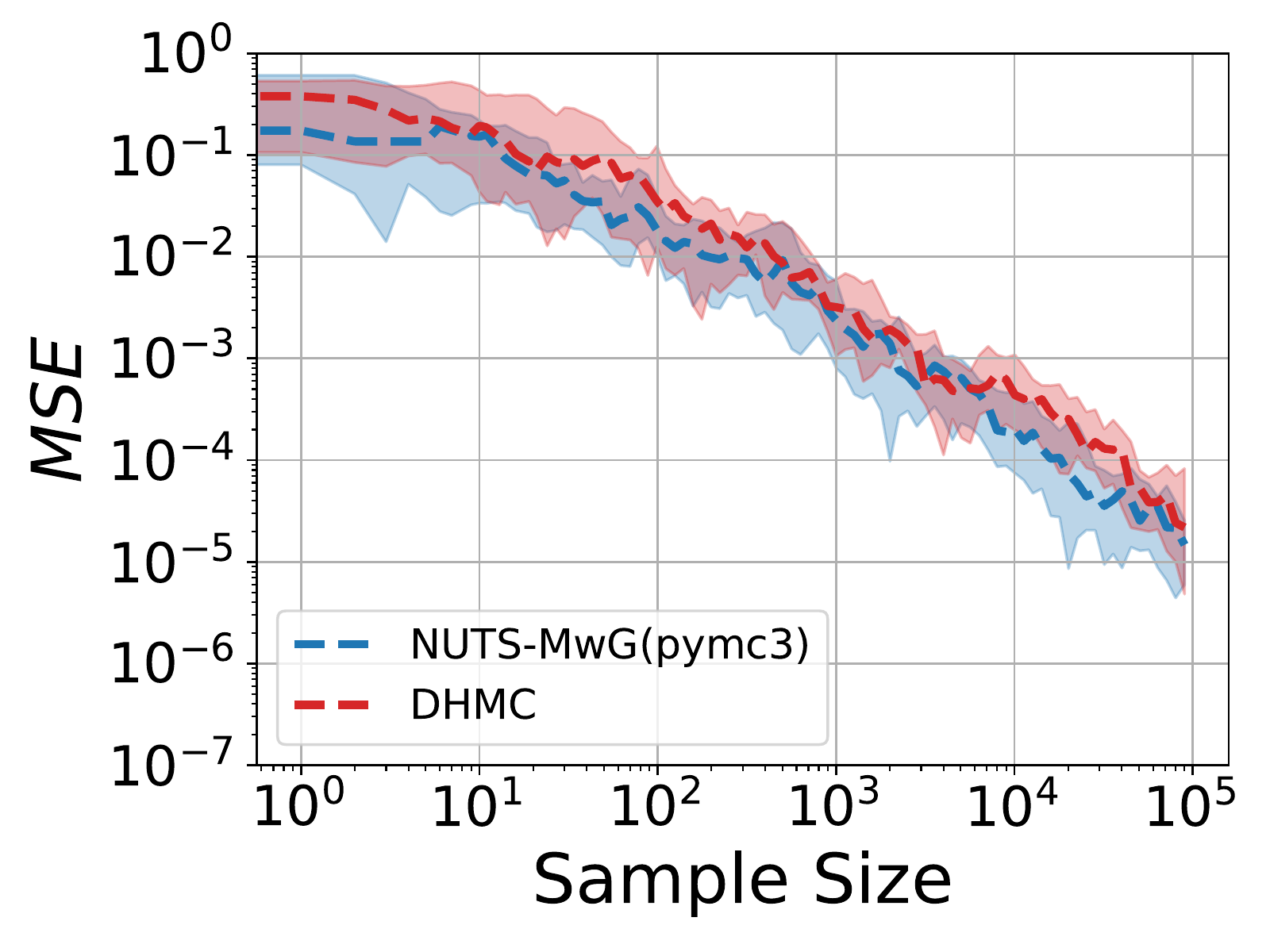}
	\vspace{-5pt}
	\caption{Mean Squared Error for the posterior estimates of the true posterior of the cluster means $\mu_{1:2}$. 
		We compare the results from our unoptimized DHMC and the optimized PyMC3 NUTS with Metropolis-within-Gibbs, and show that the performance between the two is comparable for the same computation budget.
		The median of MSE~(dashed lines) with $20\%/80\%$ confidence intervals~(shaded regions) over $20$ independent runs are plotted.}
	\label{fig:gmm-mse}
	\vspace{-10pt}
\end{figure}
\begin{figure*}[t]
	\centering		
	
	\includegraphics[width=0.184\linewidth]{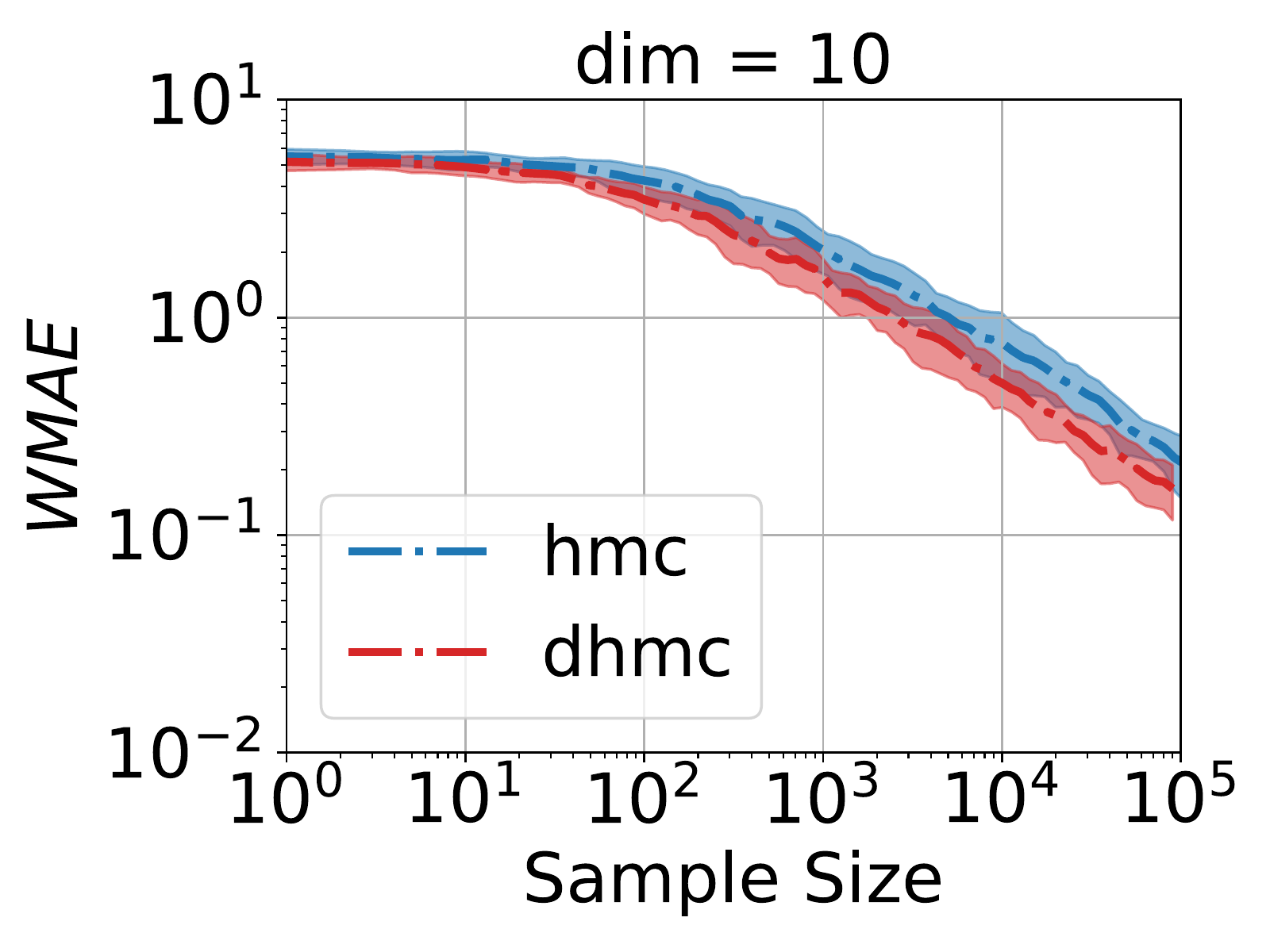}
	~
	\includegraphics[width=0.184\linewidth]{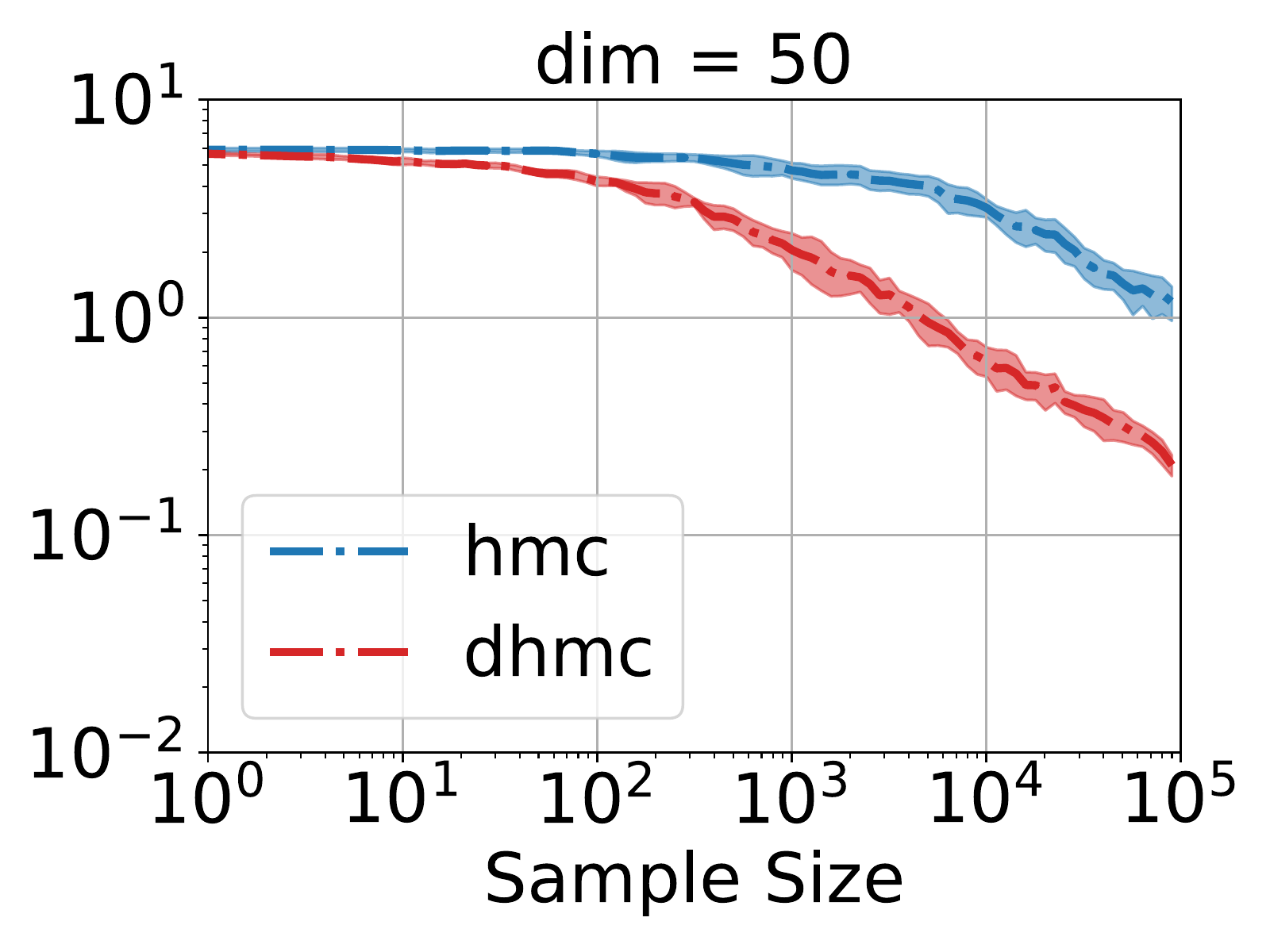}
	~
	\includegraphics[width=0.184\linewidth]{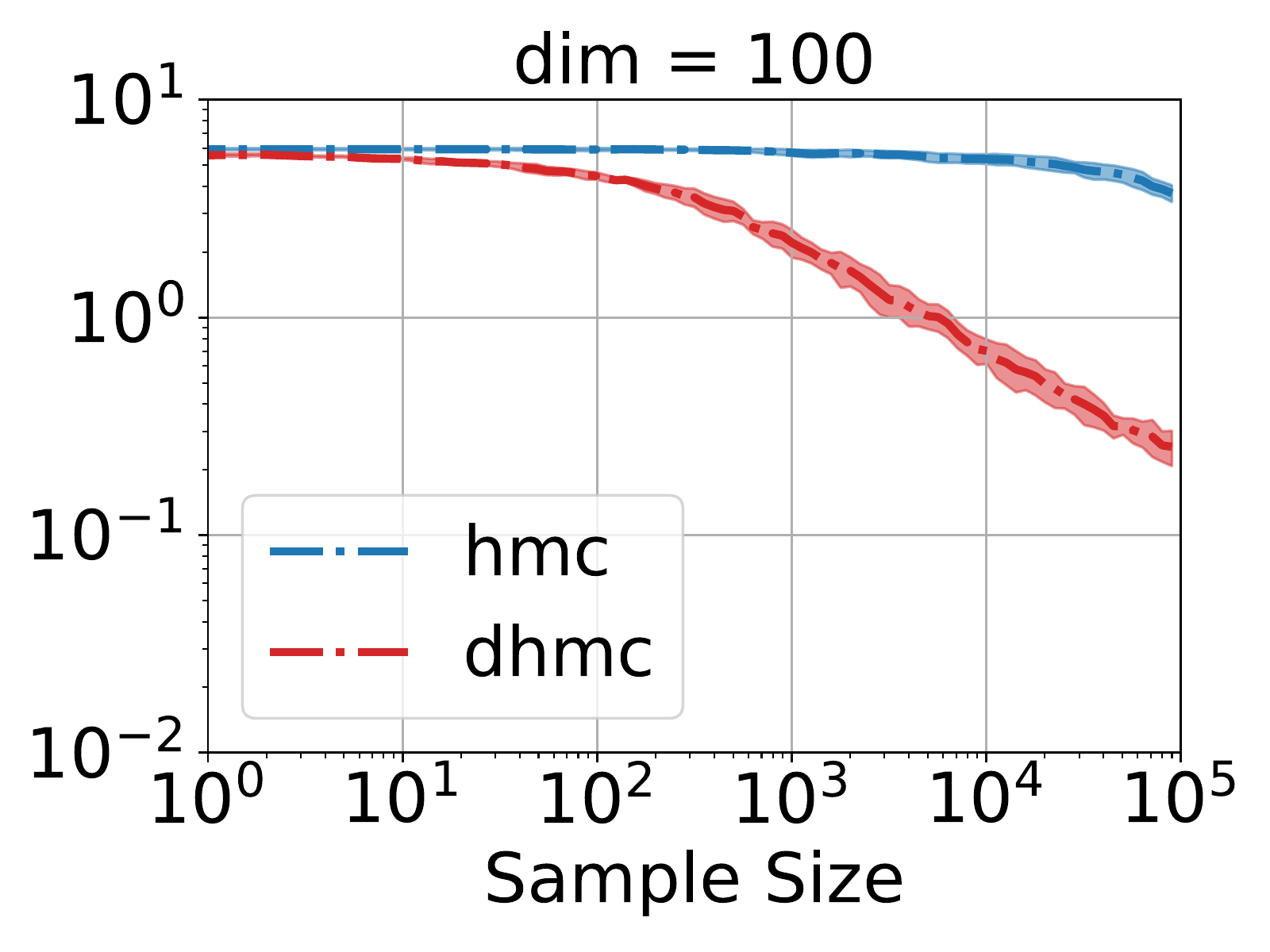}
	~
	\includegraphics[width=0.184\linewidth]{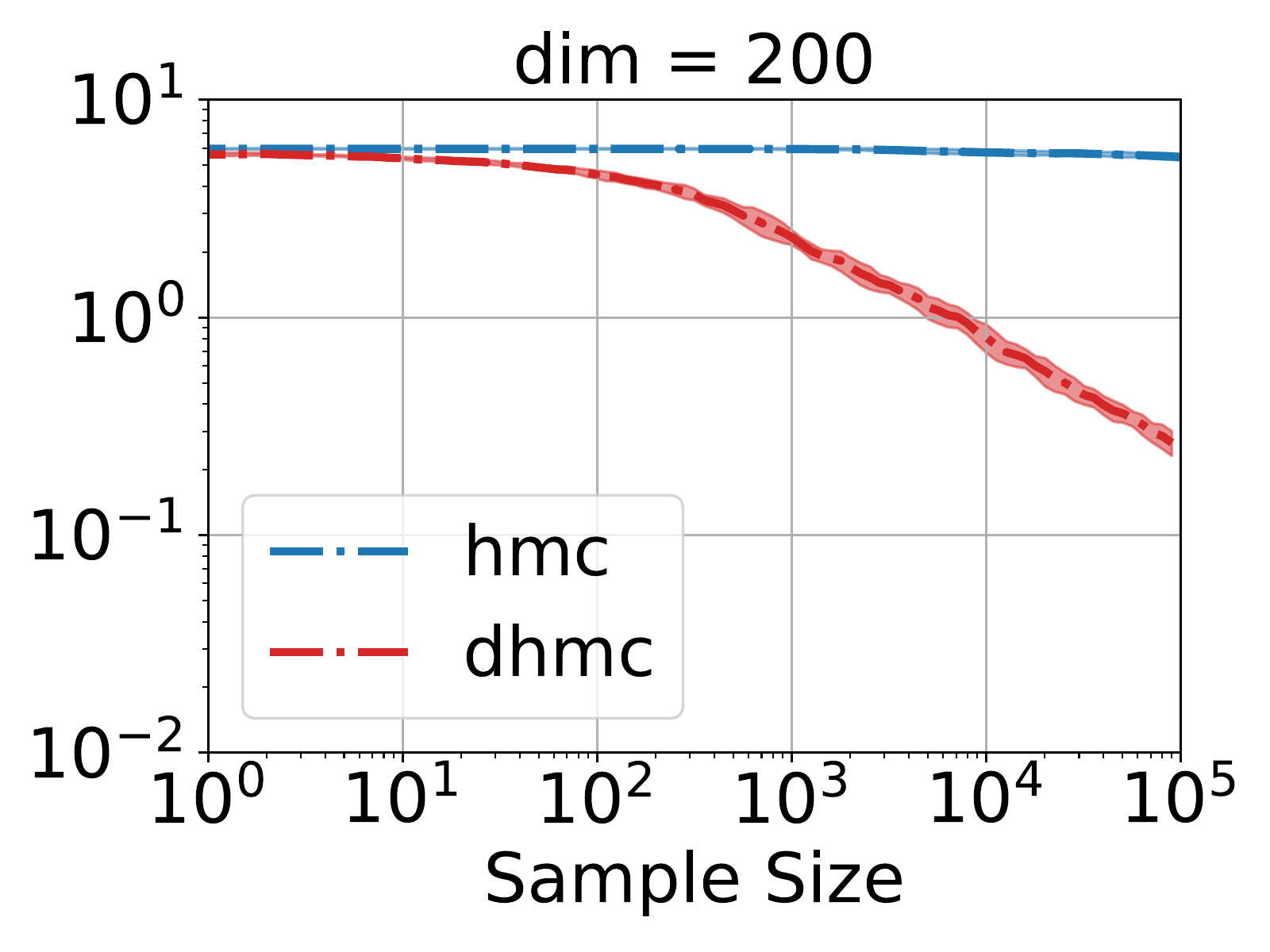}
	~
	\includegraphics[width=0.184\linewidth]{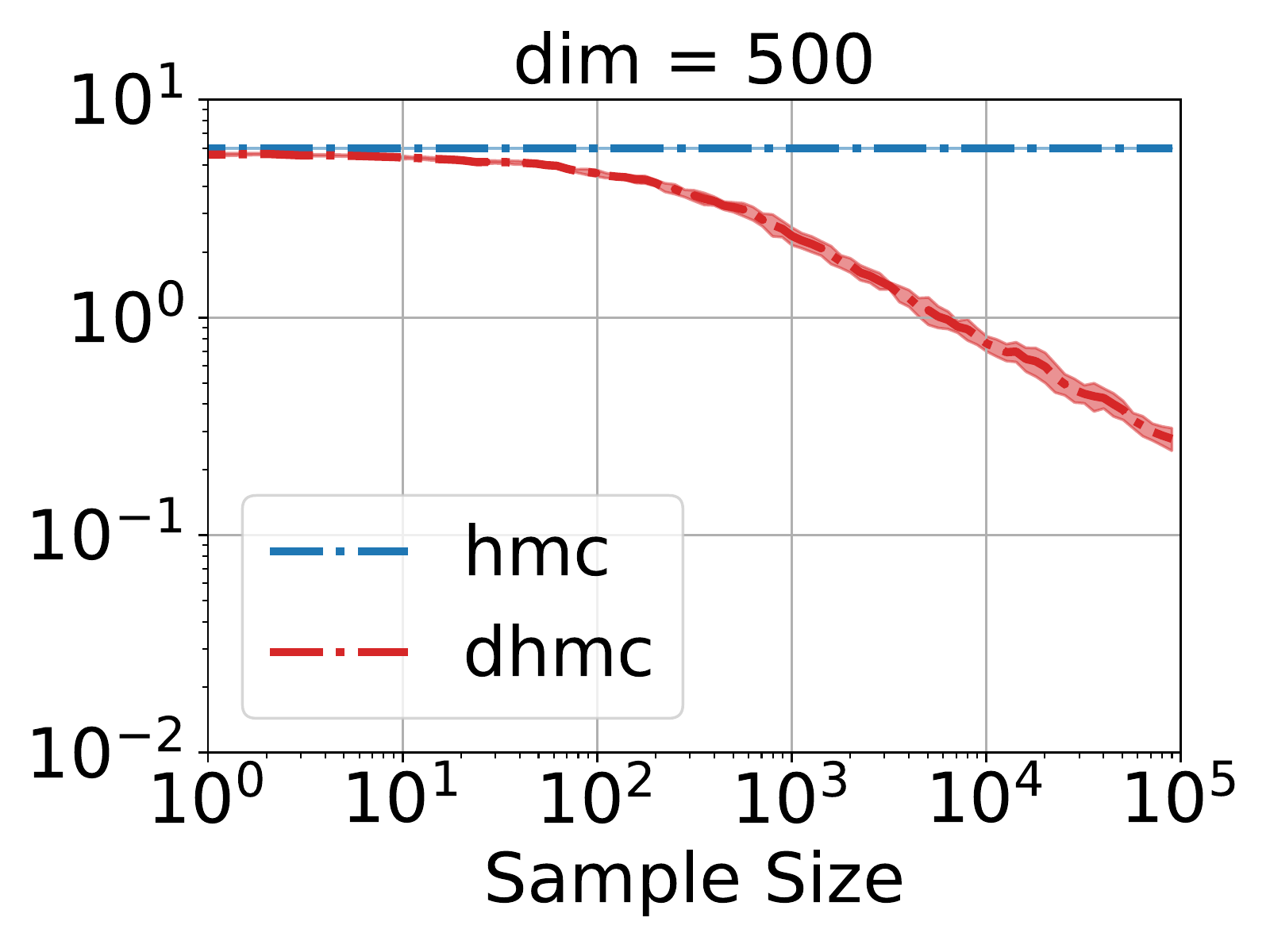}
	
	\includegraphics[width=0.184\linewidth]{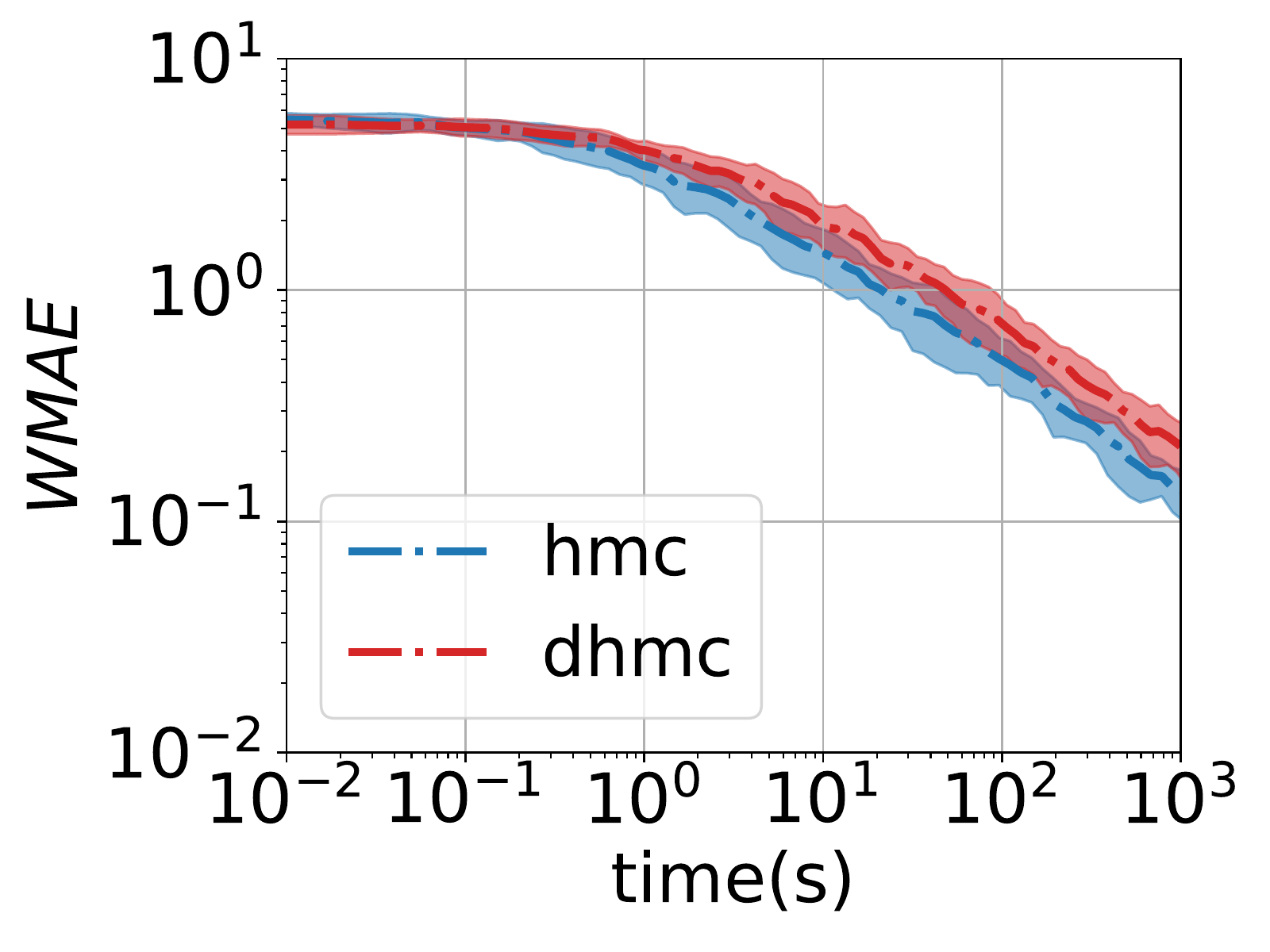}
	~
	\includegraphics[width=0.184\linewidth]{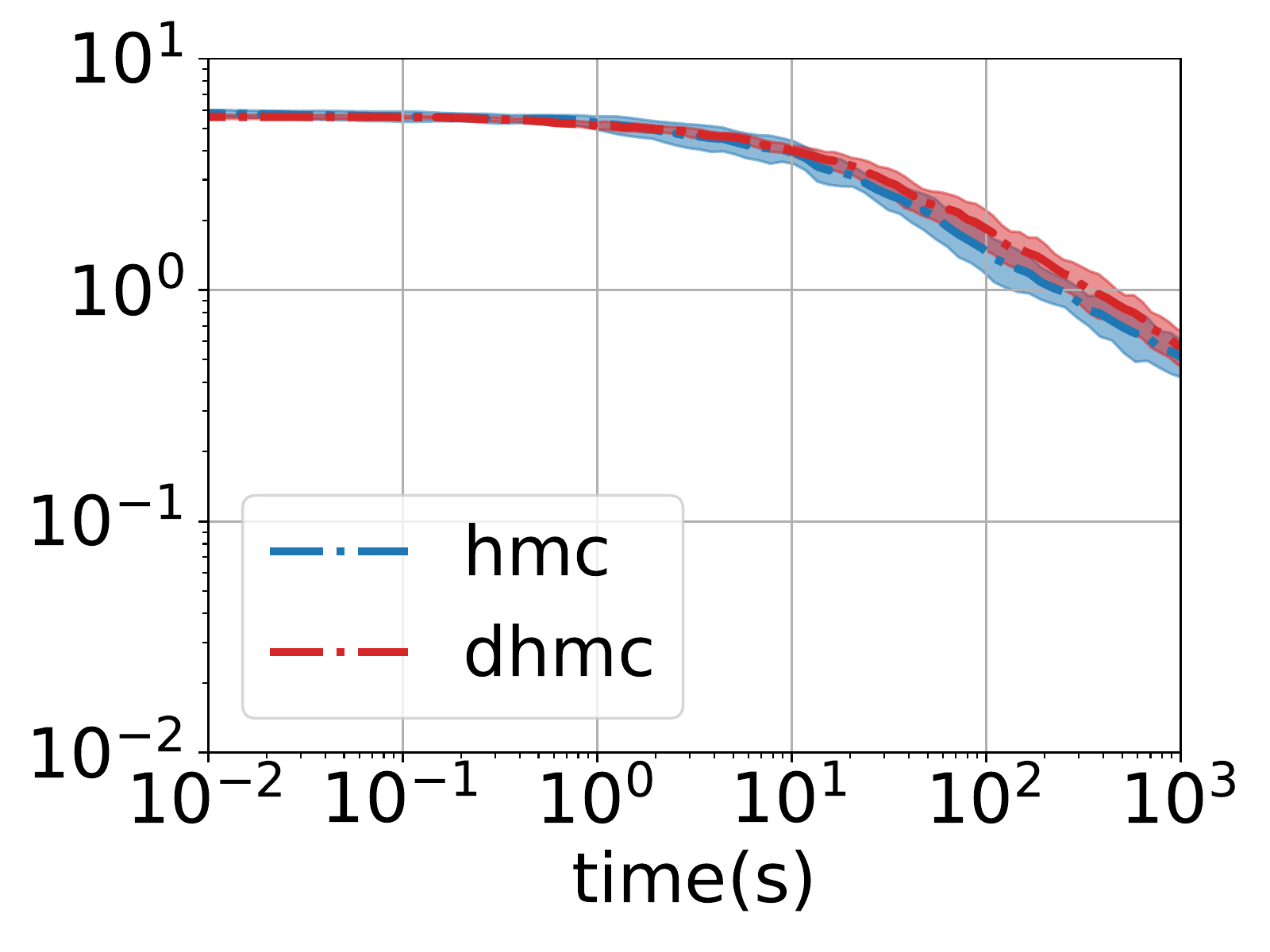}
	~
	\includegraphics[width=0.184\linewidth]{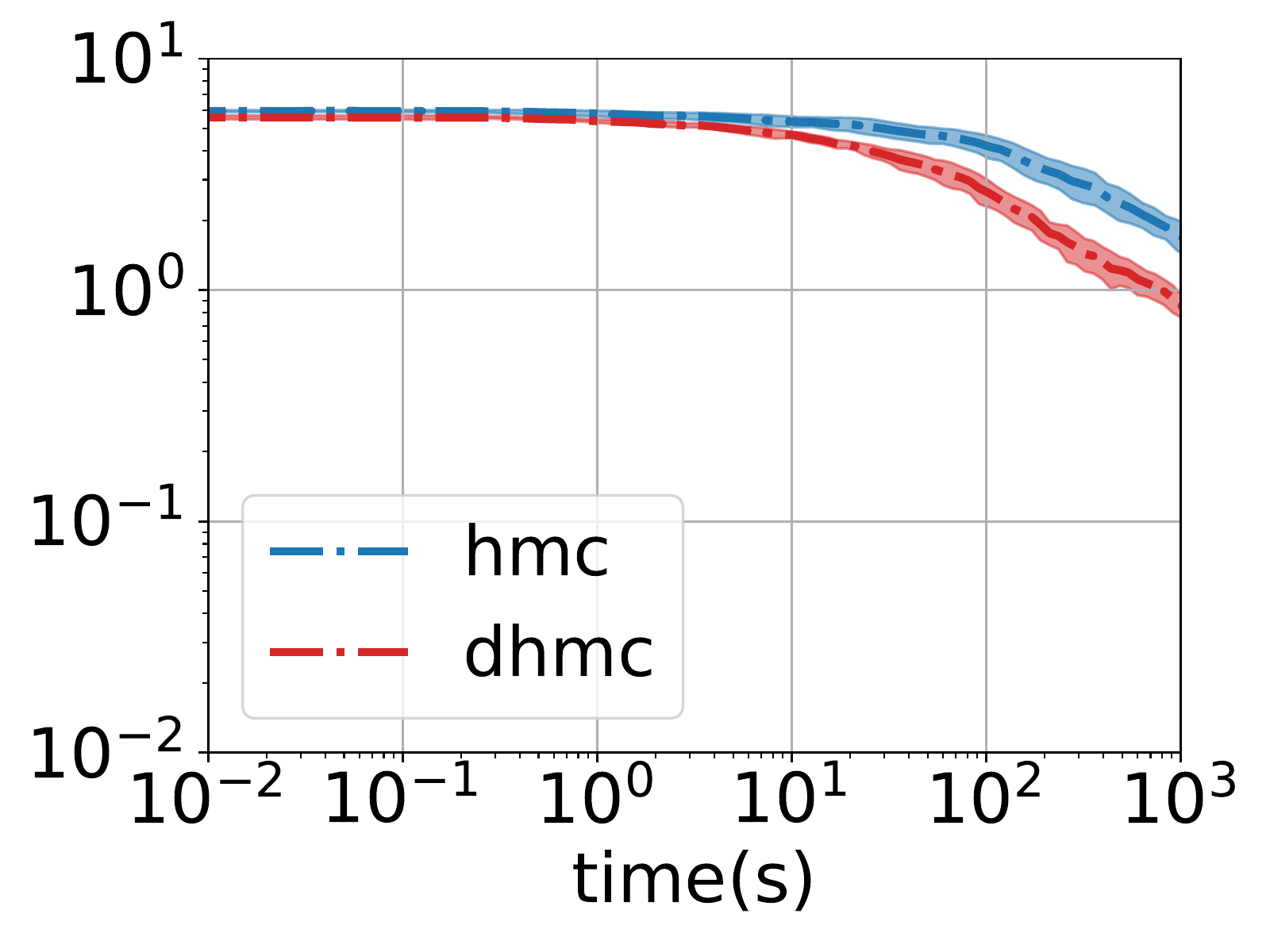}
	~
	\includegraphics[width=0.184\linewidth]{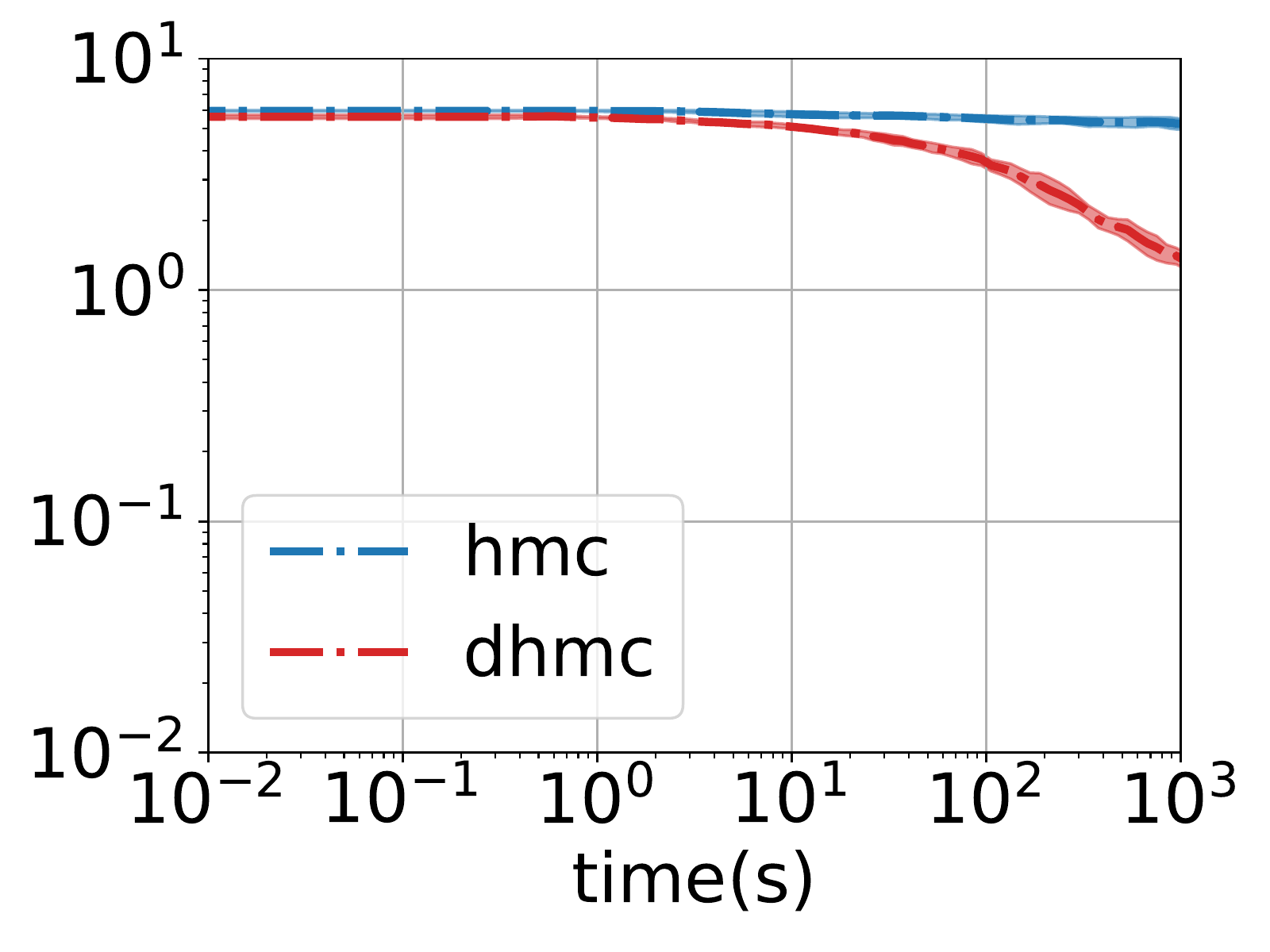}
	~
	\includegraphics[width=0.184\linewidth]{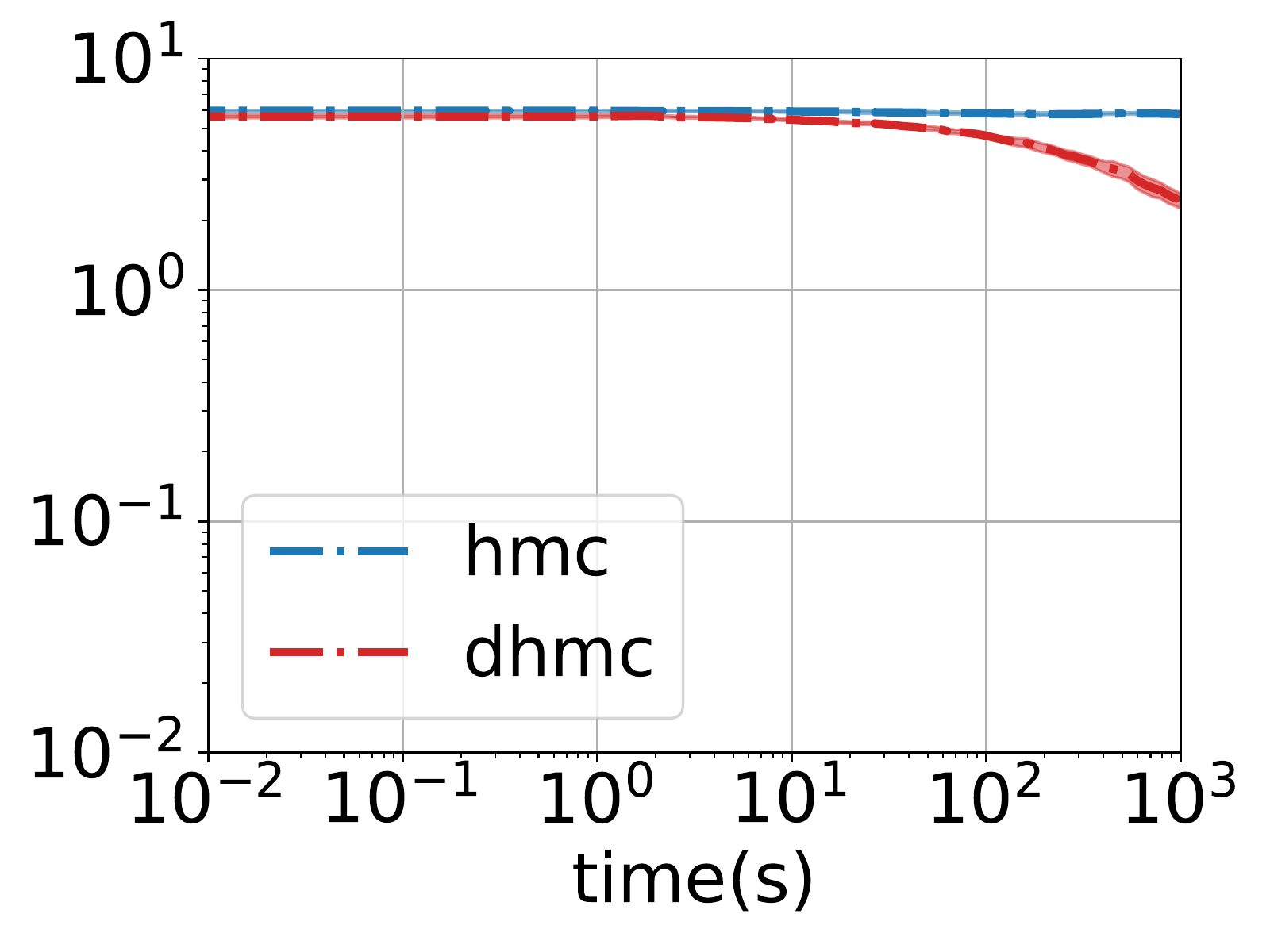}
	
	\caption{We compare DHMC against HMC on the worst mean absolute error (dashed lines) with the $20\%/80\%$ confidence intervals~(shaded regions) over 20 independent runs for dimensions $D = 10, 50, 100, 200, 500$ (left to right).
	We demonstrate how the sample efficiency decreases with respect to sample size~(\textit{top} row) and with respect to runtime~(\textit{bottom} row) respectively as dimensionality increases.
	We see that the performance of HMC deteriorates significantly more than DHMC as the dimensionality increases.
}
	\vspace{-10pt}
	\label{fig:rhmc-model}
\end{figure*}

In our first example, we demonstrate how a classic model, namely a Gaussian
mixture model, can be encoded in LF-PPL.
The density of the GMM contains a mixture of continuous and discrete
variables, where the discrete variables lead to discontinuities in the density. 
We construct the GMM as follows:
\vspace{-3pt}
\begingroup
\addtolength{\jot}{-2pt}
\begin{align*}
\mu_{k} &\sim \mathcal{N}(\mu_0, \sigma_0), \; k = 1,\dots, K\\
z_{n} & \sim \mathrm{Categorical}(p_0) , \; n = 1,\dots, N\\
y_n\, | z_n,\, & \mu_{z_n}  \sim \mathcal{N}(\mu_{z_n},\sigma_{z_n}), \; n = 1,\dots, N
\end{align*}
\endgroup
where $\mu_{1:K}, z_{1:N}$ are latent variables, $y_{1:N}$ are observed data 
with $ K$ as the number of clusters and $N$ the total number of data. 
The Categorical distribution is constructed by a combination of uniform draws and nested \lstinline[style=clojure]{if} expressions, as shown in Appendix~\ref{sec:supp-prog}. 
% In addition, one can also rewrite the program with a for-loop sugar to increase readabilty.
For our experiments, we considered a simple case with $\mu_0  \,{=}\, 0$, $\sigma_0 \,{=}\, 2$, $\sigma_{z_{1:N}}  \,{=}\,1$ and $p_0 \,{=}\, [0.5, 0.5]$, along with the
synthetic dataset:
%\[
$y_{1:N} \,{=}\, [-2.0, -2.5,  -1.7,  -1.9,  -2.2, 1.5,  2.2,  3,  1.2, 2.8]$.
%\]
We compared the Mean Squared Error~(MSE) of the posterior estimates for the cluster means of both an unoptimized version of DHMC 
and an optimized implementation of NUTS with Metropolis-within-Gibbs~(MwG) in PyMC3~\cite{salvatier2016probabilistic}, with the same computation budget.
We take $10^5$ samples and discard $10^4$ for burn in. We find that our DHMC implementation, performs comparable to the NUTS with MwG approach. 
The results are shown in Figure~\ref{fig:gmm-mse} as a function of the number of
samples.

\vspace{-2pt}
\subsection{Heavy Tail Piecewise Model}
\vspace{-3pt}
% !TEX root = main_lfppl.tex

In our next example we show how the efficiency of DHMC improves,
relative to vanilla HMC, on 
discontinuous target distributions as the dimensionality of the
problem increases.  We consider the following density\cite{afshar2015reflection} which represents a hyperbolic-like potential function, 
\begingroup
\addtolength{\jot}{-7pt}
\begin{align*}
	\pi(\boldsymbol{x}) =
	\begin{cases}
		\exp( - \sqrt{ \boldsymbol{x} ^ {T} A \boldsymbol{x} } )         & \quad \text{if }  ||  \boldsymbol{x} ||_{\infty} \leq 3  \\ 
		\exp( - \sqrt{ \boldsymbol{x} ^ {T} A \boldsymbol{x} }  - 1)    & \quad \text{if } 3 < ||  \boldsymbol{x} ||_{\infty} \leq 6  \\
		0   & \quad \text{otherwise}
	\end{cases} 
	\vspace{-10pt}
\end{align*}
\endgroup
It generates planes of discontinuities along the boundaries
defined by the \lstinline[style=clojure]{if} expressions. 
To write this as a density in our language we make use of the \lstinline[style=clojure]{factor} distribution object as shown in Appendix~\ref{sec:supp-prog}.

The results in Figure~\ref{fig:rhmc-model} provide a comparison between the DHMC and 
the standard HMC on the 
worst mean absolute error~\cite{afshar2015reflection}
as a function of the number of iterations and time, $\mathrm{WMAE}(N) = \frac{1}{N} \displaystyle\max_{d = 1, \dots, D}  \big| {\sum}_{n = 1}^{N} \boldsymbol{x}_d^{(n)} \big|$.
We see that as the dimensionality of the model 
increases, the per-sample performance of HMC deteriorates rapidly 
as seen in the top row of Figure~\ref{fig:rhmc-model}.
Even though DHMC is more expensive per iteration 
than HMC due to its sequential nature, 
in higher dimensions, the additional time costs occurred by DHMC 
is much less than the rate at which HMC
performance deteriorates.
The reason for this is that the acceptance rate of the HMC sampler degrades with
increasing dimension, while the coordinate-wise integrator of the DHMC sampler circumvents this.

\section{Conclusion}
\vspace{-5pt}
\label{sec:discussion}

In this paper we have introduced a Low-level First-order Probabilistic Programming Language (LF-PPL) and an accompanying compilation scheme for programs that have non-differentiable densities. 
We have theoretically verified the language semantics via a series of translations rules.
This ensures programs that compile in our language contain 
only discontinuities that are of measure zero.
Therefore, our language together with the compilation scheme can be used in conjunction with other
scalable inference algorithms such as adapted versions of HMC and SVI for non-differentiable densities,
 as we have demonstrated with one such variant of HMC called discontinuous HMC.
 It provides a road map for incorporating other inference algorithms into PPSs and shows the performance improvement of these inference algorithms over existing ones. 

\subsection*{Acknowledgements} 
Yuan Zhou is sponsored by China Scholarship Council (CSC). Bradley Gram-Hansen is supported by the UK EPSRC CDT in Autonomous Intelligent Machines and Systems (CDT in AIMS). 
Tom Rainforth's research leading to these results has received funding from the European Research Council under the European Union's Seventh Framework Programme (FP7/2007-2013) ERC grant agreement no. 617071.
Yang was supported by the Engineering Research Center Program through the National Research Foundation of Korea (NRF) funded by the Korean Government MSIT (NRF-2018R1A5A1059921), and also by Next-Generation Information Computing Development Program through the National Research Foundation of Korea (NRF) funded by the Ministry of Science, ICT (2017M3C4A7068177).
Kohn and Wood were supported by DARPA D3M, Intel as part of the NERSC Big Data Center, and NSERC under its Discovery grant and accelerator programs.

\bibliography{refs}
\bibliographystyle{ieeetr}

\onecolumn	
\appendix
\section{Proof of Theorem 1}
\label{sec:theorem-1-proof}
% !TEX root = main_lfppl.tex

\begin{proof} 
	Assume that $\mathcal{P}$ is piecewise smooth under analytic partition. Thus,
	\begin{equation}
	\label{eqn:good-density-HMC1}
	\mathcal{P}(x)
	=
	\sum_{i = 1}^N
	\prod_{j = 1}^{M_i} \mathbbm{1}[p_{i,j}(x) \geq 0]
	\cdot
	\prod_{l = 1}^{O_i} \mathbbm{1}[q_{i,l}(x) < 0]
	\cdot
	h_i(x)
	\end{equation}
	for some $N, M_i, O_i$ and $p_{i,j}, q_{i,l},h_i$ that satisfy the
	properties in Definition~\ref{defn:piecewise-smooth-analytic}.
	
	We use one well-known fact: the zero set $\{x \in \mathbb{R}^{n} \mid p(x) = 0\}$
	of an analytic function $p$ is the entire $\mathbb{R}^{n}$ or has zero Lebesgue measure~\cite{mityagin2015zero}. We apply the fact to each $p_{i,j}$ and deduce that
	the zero set of $p_{i,j}$ is $\mathbb{R}^{n}$ or has measure zero.
	Note that if the zero set of $p_{i,j}$ is the entire $\mathbb{R}^{n}$, the indicator
	function $\mathbbm{1}[p_{i,j} \geq 0]$ becomes the constant-$1$ function, so that it can
	be omitted from the RHS of equation \eqref{eqn:good-density-HMC1}.
	In the rest of the proof, we assume that this simplification is already done so that
	the zero set of $p_{i,j}$ has measure zero for every $i,j$.
	
	For every $1 \leq i \leq N$, we decompose the $i$-th region
	\begin{align}
	R_i = \{x \mid p_{i,j} \geq 0\ \mbox{and}\ q_{i,l}(x) < 0\ \mbox{for all $j,l$}\} \,
	\end{align}
	to
	\begin{align}
	\begin{array}{l}
	R'_i = \{x \mid p_{i,j} > 0\ \mbox{and}\ q_{i,l}(x) < 0\ \mbox{for all $j,l$}\}
	\\[0.5ex]
	R''_i = R_i \setminus R'_i.
	\end{array}
	\end{align}
	Note that $R'_i$ is open because the $p_{i,j}$ and $q_{i,l}$ are analytic and so continuous,
	both $\{r \in \mathbb{R} \mid r > 0\}$ and $\{r \in \mathbb{R} \mid r < 0\}$ are open,
	and the inverse images of open sets by continuous functions are open.
	This means that for each $x \in R'_i$, we can find an open ball at $x$ inside $R'_i$
	so that $\mathcal{P}(x') = h_i(x')$
	for all $x'$ in the ball. Since $h_i$ is smooth, this implies that $\mathcal{P}$ is differentiable
	at every $x \in R'_i$. 
	
	For the other part $R''_i$, we notice that
	\[
	R''_i \subseteq \bigcup_{j = 1}^{M_i} \{x \mid p_{i,j}(x) = 0\}.
	\]
	The RHS of this equation is a finite union of measure-zero sets, 
	so it has measure zero. Thus, $R''_i$ also has measure zero as well.
	
	Since $\{R_i\}_{1\leq i \leq N}$ is a partition of $\mathbb{R}^{n}$, we have that
	\[
	\mathbb{R}^{n} = \bigcup_{i = 1}^N R'_i \cup \bigcup_{i=1}^N R''_i.
	\]
	The density $\mathcal{P}$ is differentiable on the union of $R'_i$'s. Also, since
	the union of finitely or countably many measure-zero sets has measure zero,
	the union of $R''_i$'s has measure zero. Thus, we can set the set $A$ required
	in the theorem to be this second union.
\end{proof}

\section{Proof of Theorem 2}
\label{sec:theorem-2-proof}
% !TEX root = main_lfppl.tex

\begin{proof}
		As shown in Equation~\ref{eq:pdf}, 
		\vspace{-5pt}
		$$
		\mathcal{P} \, {:=} 
		\Big( \sum_{i = 1}^{N_D} \eta_i {\cdot} k_i \Big) \cdot \Big( \sum_{j = 1}^{N_F} \zeta_j {\cdot} l_j \Big)
		$$
        it suffices to show that both factors are non-negative and piecewise smooth under analytic partition, because such functions are closed under multiplication. 
        
        We prove a more general  result. 
        For any expression $e$, let $\mathrm{Free}(e)$ be the set of its free variables. Also, if a function $\mathcal{G}$ in Definition~\ref{defn:piecewise-smooth-analytic} 
        satisfies additionally that its $h_i$'s are analytic, we say that this function $\mathcal{G}$ is \emph{piecewise analytic} under analytic partition. We claim that for all expressions $e$ (which may contain free variables), if $e \leadsto (\Delta,\Gamma,D,F)$, 
        where 
        $D = \{(\eta_i,k_i)\,|\,1 {\leq} i {\leq} N_D\}$ 
        and $F = \{(\zeta_j,l_j, v_j)\,|\,  1 {\leq} j {\leq} N_F\}$,
        then
		$ \Big( \sum_{i = 1}^{N_D} \eta_i {\cdot} k_i \Big)$ 
		and
		$\Big( \sum_{j = 1}^{N_F} \zeta_j {\cdot} l_j \Big)$
        are non-negative functions on variables in $\mathrm{Free}(e) \cup \Delta$ and they are piecewise analytic under analytic partition,
        as $k$ and $l'$ in the sum are analytic. 
        These two properties in turn imply that 
	$\Big( \sum_{i = 1}^{N_D} \eta_i {\cdot} k_i \Big) \cdot \Big( \sum_{j = 1}^{N_F} \zeta_j {\cdot} l_j \Big)$
	is a function on variables in $\mathrm{Free}(e) \cup \Delta$
        and it is also piecewise analytic (and thus piecewise smooth) under analytic partition. Thus, the desired conclusion follows. Regarding our claim, we can prove it by induction on the structure of the expression $e$.
\end{proof}

\section{Discontinuous Hamiltonian Monte Carlo}
\label{sec:supp-dhmc}
% !TEX root = main_lfppl.tex

The discontinuous HMC (DHMC) algorithm was proposed by~\cite{nishimura2017discontinuous}. It uses a coordinate-wise integrator, Algorithm~\ref{alg:coordwise}, coupled with a Laplacian momentum to perform inference in models with non-differentiable densities.   
The algorithm works because the Laplacian momentum ensures that all discontinuous parameters move in steps of $\pm m_{b}\epsilon $ for fixed constants $m_b$ and step size $\epsilon$, where the index $b$ is associated to each discontinuous coordinate. These properties are advantages because they remove the need to know where the discontinuity boundaries between each region are; the change of the potential energy in the state before and after the $\pm m_{b} \epsilon$ move provides us with information of whether we have enough kinetic energy to move into this new region. If we do not have enough energy we reflect backwards $\mathbf{p}_{b} = -\mathbf{p}_{b}$. Otherwise, we move to this new region with a proposed coordinate update $\mathbf{x}_{b}^{*}$ and momentum $\mathbf{p}_{b} - m_{b} \cdot \mathit{sign}(\mathbf{p}_{b})\cdot \Delta U$. 
This is in contrast to algorithms such as Reflect, Refract HMC~\cite{afshar2015reflection}, that explictly need to know where the discontinuities boundaries are. Hence, it is important to have a compilation scheme that enables one to do that.

The addition of the random permutation $\phi$ of indices $b$ is to ensure that the coordinate-wise integrator satisfies the criterion of reversibility in the Hamiltonian. Although the integrator does not reproduce the exact solution, it nonetheless preserves the Hamiltonian exactly, even if the density is discontinuous. See Lemma 1 and Theorems 2-3 in \cite{nishimura2017discontinuous}. This yields a rejection-free proposal.

\begin{algorithm}[h]
	\caption{Coordinate-wise Integrator. A random permutation
		$\phi$ on $\{1,\ldots,B\}$ is appropriate if
		the induced random sequences $(\phi(1),\ldots,\phi(|B|))$ and 
		$(\phi(|B|),\ldots,\phi(1))$ have the same distribution}
	\label{alg:coordwise}
	\begin{algorithmic}[1]
		\Function{Coordinatewise($\mathbf{x},\mathbf{p},\epsilon, U$)}{}
		\State{pick an appropriate random permutation $\phi$ on $B$}
		\For{$i = 1,\ldots,B$}
		\State{$b \gets \phi(i)$}
		\State$ \mathbf{x}^{\ast} \gets \mathbf{x} $
		\State $ \mathbf{x}^{\ast}_{b} \gets \mathbf{x}_{b}^{\ast} + \epsilon m_{b} \cdot \mathit{sign}(\mathbf{p}_{b})$
		\State{$\Delta U \gets U(\mathbf{x}^{\ast}) - U(\mathbf{x})$}
		\If{$K(\mathbf{p}_{b}) = m_{b} |\mathbf{p}_{b}| > \Delta U $ }
		\State $\mathbf{x}_{b} \gets \mathbf{x}_{b}^{\ast}$
		\State $\mathbf{p}_{b} \gets \mathbf{p}_{b} - m_{b} \cdot \mathit{sign}(\mathbf{p}_{b}) \cdot \Delta U $
		\Else
		\State{$\mathbf{p}_{b} \gets - \mathbf{p}_{b} $}
		\EndIf
		\EndFor
		\State{\Return ${\mathbf{x}_{b}, \mathbf{p}_{b}}$}
		\EndFunction
	\end{algorithmic}
\end{algorithm} 

Then DHMC algorithm~\cite{nishimura2017discontinuous} adpated for LF-PPL and our compilation scheme is as follows:
\begin{algorithm}[!h]
	\label{alg:dhmcsppl1}
	\caption{Discontinuous HMC Integrator for the LF-PPL.\\
		$\chi$ is a map from random-variable names $n$ in $\Delta$ to their values $\mathbf{x}_n$, 
		$H$ is the total Hamiltonian,
		$\epsilon > 0$ is the step size, 
		and $L$ is the trajectory length.}
	\label{algo:DHMC-SPPL}
	\begin{algorithmic}[1]
		\Function{DHMC-LFPPL}{$\Delta,\Gamma,D,F, \mathbf{x},\mathbf{p}, H,\epsilon, L $}
		\State{$B = \Gamma$;$\ \ $ $A = \Delta \setminus \Gamma$}
		
		\For{$a \in A$} \Comment{$a$ represents the set of continuous variables}
		\State{$\mathbf{x}^{0}_a \gets \mathbf{x}_a$;$\ \ $ $ \mathbf{p}_{a} \sim \mathcal{N}(\mathbf{0}, \mathbf{1})$}
		\EndFor
		\For{$b \in B$}
		\State{$\mathbf{x}^{0}_b \gets  \mathbf{x}_b$;$\ \ $ $ \mathbf{p}_b \sim \mathit{Laplace}(\mathbf{0}, \mathbf{1})$}\Comment{$b$ represents the set of discontinuous variables}
		\EndFor
		\State{$\forall a \in A$, $\, \mathbf{x}^{0}_a \gets \mathbf{x}_a$;$\ \ $ $ \mathbf{p}_{a} \sim \mathcal{N}(\mathbf{0}, \mathbf{1})$} \Comment{$A$ represents the set of continuous variables}
		\State{$\forall b \in B$, $\, \mathbf{x}^{0}_b \gets \, \mathbf{x}_b$;$\ \ $ $ \mathbf{p}_b \sim \mathit{Laplace}(\mathbf{0}, \mathbf{1})$}\Comment{$B$ represents the set of discontinuous variables}
		
		\State{$U \gets -\mbox{\sc LogJointDensity}(D,F)$}
		\For{$i = 1$ to $L$}
				\State{$U_A \gets U\ \mbox{with names in $B$ replaced by their values in $\mathbf{x}^i_B$}$}
		\State{$(\mathbf{x}^{i}_{A}, \mathbf{p}^{i}_{A}) \gets ${\sc Halfstep1}$(\mathbf{x}^{i-1}_{A},\mathbf{p}^{i-1}_A, \epsilon, U_A)$ }
				\State{$U_B \gets U\ \mbox{with names in $A$ replaced by their values in $\mathbf{x}^i_A$}$}
		\State{$(\mathbf{x}^{i}_B,\mathbf{p}^{i}_B)\gets ${\sc Coordinate-wise}($\mathbf{x}^{i-1}_{B}, \mathbf{p}^{i-1}_{B},\epsilon, U_B$)}
				\State{$U_A \gets U\ \mbox{with names in $B$ replaced by their values in $\mathbf{x}^i_B$}$}
		\State{$(\mathbf{x}^{i}_{A}, \mathbf{p}^{i}_{A}) \gets ${\sc Halfstep2}$(\mathbf{x}^i_A, \mathbf{p}^i_A, \epsilon, U_A)$}
		\EndFor
		\State{$\mathbf{x}^{L} \gets \mathbf{x}^{L}_{A} \cup \mathbf{x}^{L}_{B}$,  $\; \mathbf{p}^{L} \gets \mathbf{p}^{L}_{A} \cup \mathbf{p}^{L}_{B}$};
		\State{$\mathbf{x}^{*}, \mathbf{p}^{*} \gets \mbox{\sc Evaluate}(F, \; \mathbf{x}^L, \mathbf{p}^{L})$}
		\State{ $\alpha \sim Uniform(0,1)$}
		\If{ $\alpha > \min\{1,\exp(H(\mathbf{x},\mathbf{p})-H(\mathbf{x}^{*},\mathbf{p}^{*}))\}$}
		\State{\Return$\mathbf{x}^{*}, \mathbf{p}^{*}$}
		\Else
		\State{\Return$\mathbf{x}, \mathbf{p}$}
		\EndIf
		\EndFunction
				\Function{HALFSTEP1}{$\mathbf{x}, \mathbf{p}, \epsilon, U$}
				\State {$\mathbf{p}' \gets \mathbf{p} - \frac{\epsilon}{2} \nabla_{\mathbf{x}} U(\mathbf{x})$}
				\State {$\mathbf{x}' \gets \mathbf{x} + \frac{\epsilon}{2}\nabla_{\mathbf{p}'} K(\mathbf{p}')$}
				\State {\Return ($\mathbf{x}', \mathbf{p}'$)}
				\EndFunction
				\Function{HALFSTEP2}{$\mathbf{x}, \mathbf{p}, \epsilon, U$}
				\State {$\mathbf{x}' \gets \mathbf{x} + \frac{\epsilon}{2}\nabla_{\mathbf{p}} K(\mathbf{p})$}
				\State {$\mathbf{p}' \gets \mathbf{p} - \frac{\epsilon}{2} \nabla_{\mathbf{x'}} U(\mathbf{x'})$}
				\State {\Return ($\mathbf{x}', \mathbf{p}'$)}
				\EndFunction

	\end{algorithmic}
\end{algorithm}

\clearpage
\section{Program code}
\label{sec:supp-prog}
% !TEX root = main_lfppl.tex

\begin{figure}[h!]
\centering
\begin{minipage}{.5\textwidth}
  \centering
  \label{fig:gmm1a}
\begin{lstlisting}[basicstyle=\ttfamily\scriptsize,style=default]
(let [y (vector -2.0  -2.5  ... 2.8)
      pi [0.5 0.5]
      z1 (sample (categorical pi))
      ... 
      z10(sample (categorical pi))
      mu1 (sample (normal 0 2))
      mu2 (sample (normal 0 2))
      mus (vector mu1 mu2)]
  (if (< (- z1) 0)                 
       (observe (normal mu1 1) (nth y 0))
       (observe (normal mu2 1) (nth y 0)))
  ...
  (if (< (- z10) 0)
      (observe (normal mu1 1) (nth y 9))
      (observe (normal mu2 1) (nth y 9)))
  (mu1 mu2 z1 ... z10))
\end{lstlisting}
  \captionof{figure}{The LF-PPL version of the Gaussian mixture model detailed in Section~\ref{sec:dhmcLF-PPL}.}
\end{minipage}%
\begin{minipage}{.5\textwidth}
  \centering
  \begin{lstlisting}[basicstyle=\scriptsize,style=default]
(let [x (sample (uniform -6 6))
       abs-x (max x (- x))
       z (- (sqrt (* x (* A x))))]
   (if (< (- abs-x 3) 0)
       (observe (factor z) 0)
       (observe (factor (- z 1)) 0))
  x)
  \end{lstlisting}
  \vspace{-10pt}
  \label{fig:heavy-tail-LF-PPL}
  \captionof{figure}{The LF-PPL version of the heavy-tailed model detailed in Section~\ref{sec:dhmcLF-PPL}.}
  \label{fig:test2}
\end{minipage}
\end{figure}

\end{document}